\documentclass[11pt]{article}

\usepackage[final]{acl}

\usepackage{times}
\usepackage{latexsym}

\usepackage[T1]{fontenc}

\usepackage[utf8]{inputenc}

\usepackage{microtype}

\usepackage{inconsolata}

\usepackage{graphicx}

\usepackage{url}            
\usepackage{booktabs}       
\usepackage{amsfonts}       
\usepackage{nicefrac}       
\usepackage{microtype}      
\usepackage{xcolor}         

\usepackage{array}
\usepackage{multirow}
\usepackage{color}
\usepackage{amsmath}
\usepackage{xcolor}
\usepackage{listings, xcolor}
\usepackage{colortbl}
\usepackage{xspace}
\usepackage{booktabs}
\usepackage{bbding}
\usepackage{graphicx}
\usepackage{enumitem}
\usepackage{booktabs}
\usepackage{bbding}
\usepackage{graphicx} 
\usepackage{wrapfig}
\usepackage{tcolorbox}
\usepackage{listings}
\tcbuselibrary{skins}
\usepackage{tabularx}
\usepackage{makecell}
\usepackage{amssymb}
\usepackage{pifont}
\usepackage{float}
\usepackage{bbding}
\usepackage{fancyvrb}
\usepackage{tcolorbox}
\usepackage{lipsum}
\usepackage{caption}
\usepackage{float}
\usepackage{subcaption}
\usepackage{placeins}
\usepackage{etoc}
\usepackage{makecell}
\usepackage{color, xspace}
\usepackage[dvipsnames]{xcolor}
\newcommand{\ourmodel}{{\texttt{\textbf{MemCoE}}}\xspace}
\newcommand{\moduleone}{{\texttt{\textbf{MGI}}}\xspace}
\newcommand{\moduletwo}{{\texttt{\textbf{GMPO}}}\xspace}

\definecolor{PrefrontalBlue}{RGB}{61,114,181}
\definecolor{HippoOrange}{RGB}{191,93,27}

\newcolumntype{M}[1]{>{\centering\arraybackslash}m{#1}}

\title{Learning How and What to Memorize: Cognition-Inspired Two-Stage Optimization for Evolving Memory }

\author{Derong Xu\textsuperscript{1,2}, Shuochen Liu\textsuperscript{1}, Pengfei Luo\textsuperscript{1}, Pengyue Jia\textsuperscript{2}, Yingyi Zhang\textsuperscript{2,3},\\ 
{\bf Yi Wen\textsuperscript{2}, Yimin Deng\textsuperscript{2,4}, Wenlin Zhang\textsuperscript{2}, Enhong Chen\textsuperscript{1}, Xiangyu Zhao\textsuperscript{2,}\thanks{Corresponding authors.}, Tong Xu\textsuperscript{1,}\footnotemark[1]}\\
        \textsuperscript{1}University of Science and Technology of China \& State Key Laboratory of\\ Cognitive Intelligence,
        \textsuperscript{2}City University of Hong Kong \\
        \textsuperscript{3}Dalian University of Technology, 
        \textsuperscript{4}Xi'an Jiaotong University \\
derongxu@mail.ustc.edu.cn, xianzhao@cityu.edu.hk, tongxu@ustc.edu.cn
}

\begin{document}
\maketitle
\begin{abstract}
Large language model (LLM) agents require long-term user memory for consistent personalization, but limited context windows hinder tracking evolving preferences over long interactions. Existing memory systems mainly rely on static, hand-crafted update rules; although reinforcement learning (RL)-based agents learn memory updates, sparse outcome rewards provide weak supervision, resulting in unstable long-horizon optimization. Drawing on memory schema theory and the functional division between \emph{prefrontal regions} and \emph{hippocampus regions}, we introduce MemCoE, a cognition-inspired two-stage optimization framework that learns \textbf{\texttt{how}} memory should be organized and \textbf{\texttt{what}} information to update. In the first stage, we propose \textbf{Memory Guideline Induction} to optimize a global guideline via contrastive feedback interpreted as textual gradients; in the second stage, \textbf{Guideline-Aligned Memory Policy Optimization} uses the induced guideline to define structured process rewards and performs multi-turn RL to learn a guideline-following memory evolution policy. We evaluate on three personalization memory benchmarks, covering explicit/implicit preference and different sizes and noise, and observe consistent improvements over strong baselines with favorable \textbf{robustness}, \textbf{transferability}, and \textbf{efficiency}\footnote{https://github.com/Applied-Machine-Learning-Lab/ACL2026\_MemCoE}.

\end{abstract}

\section{Introduction}

\begin{figure}[t]
\centering
\includegraphics[width=0.99\columnwidth]{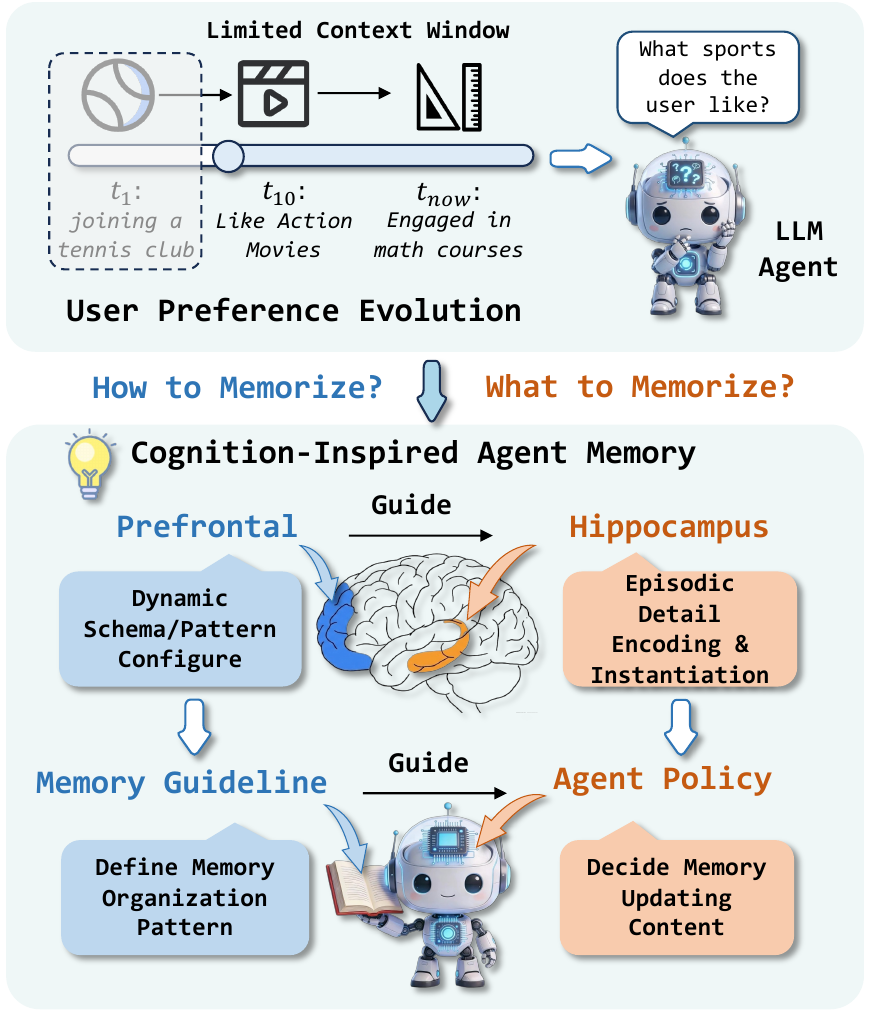}
\caption{ Top: With a limited context window, the agent fails to capture preferences. Bottom: Inspired by \textcolor{PrefrontalBlue}{\textbf{\texttt{Prefrontal}}} $\rightarrow$ \textcolor{HippoOrange}{\textbf{\texttt{Hippocampus}}}, we decouple agent memory into \textcolor{PrefrontalBlue}{\textbf{\texttt{Memory Guideline}}} (for organizing) $\rightarrow$ \textcolor{HippoOrange}{\textbf{\texttt{Agent Memory}}} (for updating).}
\label{fig:intro}
\end{figure}

Large language models (LLMs) have demonstrated remarkable capabilities as conversational agents in various real-world applications, such as personal assistant \cite{li2024personalsurvey}, customer service \cite{RMM,zhang2025notellm}, and education \cite{AI4EDU}.
In these settings, adaptive personalized interaction depends on the agent's ability to continuously integrate information about a user’s evolving preferences and habits \cite{personamem,jiang2025personamemv2}.
However, the context window prevents LLMs from retaining and exploiting the full history of dialogue \cite{parallelencoding}, and simply storing and retrieving past dialogue snippets struggles to capture such dynamic patterns \cite{li2025memos}.
This limitation highlights the necessity of maintaining an external memory that preserves important information over time, enabling consistent and personalized responses \cite{chhikara2025mem0,memorybank,deng2026enhancing}.

Many existing LLM agent memory systems typically build a workflow that converts raw dialogue into external memory banks \cite{xu2026from,xu2025amem,fang2025lightmem}. Yet these methods rely on static and predefined pipelines for extraction rules, making it difficult to learn from interaction feedback or adapt to user behavior. To solve this, other works treat memory operations as learnable actions and train an end-to-end memory update policy with reinforcement learning (RL) \cite{yu2025memagent,memory-r1,wang2025memalpha}. While more adaptive, memory updates typically involve free-form edits over what to write or forget. When guided by only simple instructions and optimized with sparse and delayed outcome-level rewards, the policy is weakly constrained and faces a large action space, making exploration and long-horizon optimization challenging. This often results in unstable training and increased data requirements \cite{agentrlsurvey}, motivating the need for more effective mechanisms for memory organization and updating.

Memory schema theory \cite{alba1983memoryschematic} in cognitive psychology, as shown in Figure~\ref{fig:intro} bottom, offers a perspective for understanding how human memory is organized and updated. Specifically, the theory suggests a functional division of labor between brain systems: \textbf{Prefrontal regions} dynamically select and configure an appropriate schema based on the current context, thereby shaping expectations and attentional priorities, while the \textbf{Hippocampus regions} \cite{teyler1986hippocampal} instantiate this schema by encoding the concrete episodic details of ongoing experience. 
Importantly, this division is advantageous because it maintains a \textbf{stable schema-level organizing prior} that guides attention and structuring, while allowing the hippocampus system to flexibly encode context-specific episodic details within that scaffold. From this perspective, the separation naturally decouples how memory is controlled (i.e., the organization patterns) and what is stored (i.e., the update content).
Motivated by this mechanism, we ask the following question:
\definecolor{cvprblue}{rgb}{0.21,0.49,0.74}
\begin{tcolorbox}[colframe=black!50, colback=cvprblue!8, boxrule=1.5pt, arc=2mm, top=4pt, bottom=4pt, left=4pt, right=4pt,  boxsep=1pt]
\raisebox{-0.2\baselineskip}{\includegraphics[height=1\baselineskip]{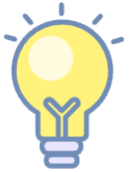}}
\textit{Can we build a schema-based agent memory system whose organization and updating mechanisms evolve in a manner analogous to human brain?}
\end{tcolorbox}

To answer this question, in this paper, we introduce \ourmodel, a two-stage optimization framework inspired by a functional analogy to human memory, enabling the agent to learn \textbf{\texttt{how}} memory should be organized, and \textbf{\texttt{what}} content should be stored and updated, by optimizing a memory guideline and a policy for evolving memory. Our approach maintains a user memory bank that evolves alongside the dialogue. In the first stage, to simulate the Prefrontal regions, we propose $\clubsuit\ $\textbf{Memory Guideline Induction (MGI)}, which treats the instruction prompt as a global natural-language parameter and optimizes it via two key techniques: (i) interpreting contrastive feedback over memory-augmented trajectories as textual gradients and (ii) aggregating these gradients at the batch level, thereby inducing a domain-agnostic textual guideline. In the second stage, we propose $\spadesuit\ $\textbf{Guideline-Aligned Memory Policy Optimization (GMPO)}, which encodes instance-specific episodic details via guideline-aligned policy optimization. GMPO utilizes the optimized guideline to define guideline-aligned rewards and performs multi-turn RL over memory-augmented trajectories to train the memory-evolution policy end-to-end, thereby jointly encouraging adherence to the guideline and determining what content should be memorized. Crucially, the first stage induces a guideline that defines a stable set of memory operations, effectively constraining the action space explored by the policy. Given this constrained space, the second stage can focus on process-level guideline reward, learning to invoke the guideline-specified operations with appropriate content.

We empirically evaluate \ourmodel on three personalization memory benchmarks (PersonaMem, PrefEval, and PersonaBench), spanning explicit and implicit preference and increasingly noisy evidence sources, where accurate answering requires tracking evolving user states over extended histories. Across these settings, \ourmodel consistently outperforms strong baselines built on static memory templates or RL-based memory updating, while remaining \textbf{efficient} in memory evolution and \textbf{scalable} to longer contexts and more rounds. Moreover, the induced guideline exhibits strong \textbf{transferability} across LLMs, supporting \textbf{robust generalization} under distribution shifts in query type.


\section{Related Work}

\paragraph{Memory for LLM Agents.}
Memory has become a foundational capability for LLM agents, supporting long-horizon understanding, continual adaptation in complex environments\cite{zhang2024survey_memoery,wu2025humansurveymemory,hu2025memoryageaiagentssurvey}.
To equip LLM agents with memory, most methods construct an explicit memory bank supported by primitives for  segmentation \cite{secom}, summarization \cite{kim2024theanine,RecurSum,SumMem,Think-in-memory,memochat,rasmussen2025zep}, compression \cite{chen2025compress,ReadAgent,recomp,COMEDY}, and forgetting/updating to maintain long-term quality \cite{memorybank,LD-Agent}.
To improve retrieval quality, several approaches build structured memory indices such as trees \cite{memtree,raptor} and graphs \cite{hipporag2,chhikara2025mem0,wang2025mirix,xu-etal-2024-multi}. Beyond generic storage, personalization-oriented methods emphasize capturing user profiles and preferences for downstream conditioning \cite{zhang2026personalize,xu2025harnessing,zhang2025personaagent,perltqa,fang2025lightmem,memorag}. Another thread focuses on experience memory, where agents memorize successful or failed trajectories to improve future decision-making \cite{packer2023memgpt,AWM,tang2025agentkb,ouyang2025reasoningbank,zhao2024expel,zhang2025gmemory,gao2025llm4rerank,jia2024mill}.  Despite strong performance, these methods rely on hand-crafted heuristics, which can be brittle under non-stationary user behaviors. Our method is orthogonal to memory construction and storage, and is broadly compatible with existing memory backends as a general mechanism for memory evolution to support long-term personalization.

\paragraph{RL for Memory.}
Recent works treat memory operations as a sequential decision problem and optimize it with reinforcement learning~\cite{zhang2026evoking,wang2025memalpha,zhou2025mem1,memory-r1,m3-agent,yuan2025memsearcher,zhang2025MemoryasAction,liu2025large,li2025towards}.
For example, RMM~\cite{RMM} learns to manage long-term personalized memory via reflective update and retrieval; MemAgent~\cite{yu2025memagent} uses RL to learn a memory agent that maintains a fixed-length context by selectively preserving/overwriting long dialogue history; MEM1~\cite{zhou2025mem1} trains memory and reasoning synergy to form compact memory for efficient long-horizon agent; MemGen~\cite{zhang2025memgen} proposes generative latent memory that weaves experience into reusable memory tokens for self-evolving agents.
However, these approaches typically rely on final success/answer as sparse rewards, and lack process-level rewards that directly guide how memory should be updated. We address this by introducing \emph{guideline-aligned rewards}, which provide structured learning signals for memory evolution.

\paragraph{Prompt Optimization.}
A growing line of work treats prompts as optimizable natural-language parameters, iteratively refining them via model-generated feedback rather than numerical gradients~\cite{yuksekgonul2025optimizing,yang2023large,pryzant2023automatic,shinn2023reflexion,tang2025unleashing,zhang2024offline,zhang2024revolve,zhang2024aflow}.
The common pattern involves evaluating the current prompt, generating natural-language edit signals (textual gradients), and applying them to produce improved variants.
For instance, TextGrad~\cite{yuksekgonul2025optimizing} uses LLM-generated textual gradients for iterative prompt refinement; OPRO~\cite{yang2023large} treats the LLM as a black-box optimizer that proposes and scores prompt candidates; Reflexion~\cite{shinn2023reflexion} converts past failures into self-reflection feedback carried forward to guide future actions.
While related in spirit, MGI differs in that it optimizes a global memory-evolution guideline over long histories rather than a single-turn prompt, stabilizes updates via contrastive diagnosis and batch aggregation, and interfaces with Stage-2 RL through guideline-aligned process rewards to jointly optimize memory update policies and reinforced behaviors.
\section{Preliminary}
We consider a conversational setting in which a user interacts with an assistant agent over time. Let \( h_t \) denote the \( t \)-th dialogue snippet, and let the cumulative interaction history be represented as a dialog set
\(
\mathcal{H} = \{h_1, h_2, \ldots, h_t\}.
\)
To support long-term personalization, the system maintains a \textbf{user memory bank} \( \mathcal{M}_t \), a textual representation that evolves dynamically as new user behaviors and preferences emerge.
Beyond dialogue content \( h_t \), we incorporate a learnable \emph{memory-update prompt} \( \mathcal{S} \) as a parameterized system component that regulates how memory is updated.
Formally, the overall memory bank-evolution process is summarized in generic form with an evolution module \( \mathcal{T} \):
\begin{equation}
\mathcal{M}_{t+1} = \mathcal{T}(\mathcal{M}_t, h_t ; \mathcal{S}, \phi),
\end{equation}
where $\phi$ denotes the parameters of LLM. The evolution operator \( \mathcal{T} \) encapsulates the mechanisms for incorporating new information, refining existing entries, and removing outdated or inconsistent content. This formulation treats user memory as a continuously adapting latent structure aligned with the user’s evolving profile.

Given a task or query input \( x \), the agent \( \mathcal{A} \) generates a personalized response conditioned on \( x \) and the current memory state \( \mathcal{M}_t \):
\(
y_t = \mathcal{A}(x, \mathcal{M}_t),
\)
indicating that \( \mathcal{M}_t \) serves as auxiliary context modulating the agent’s behavior. The central challenge in our task, therefore, lies in designing a principled mechanism  \( \mathcal{T} \) that allows \( \mathcal{M}_t \) to evolve coherently with \( \mathcal{H} \), enabling the agent to maintain stable, accurate, and temporally consistent user representations throughout long-term interaction.

\section{Methodology}
\begin{figure*}
\centering
\includegraphics[width=0.99\linewidth]{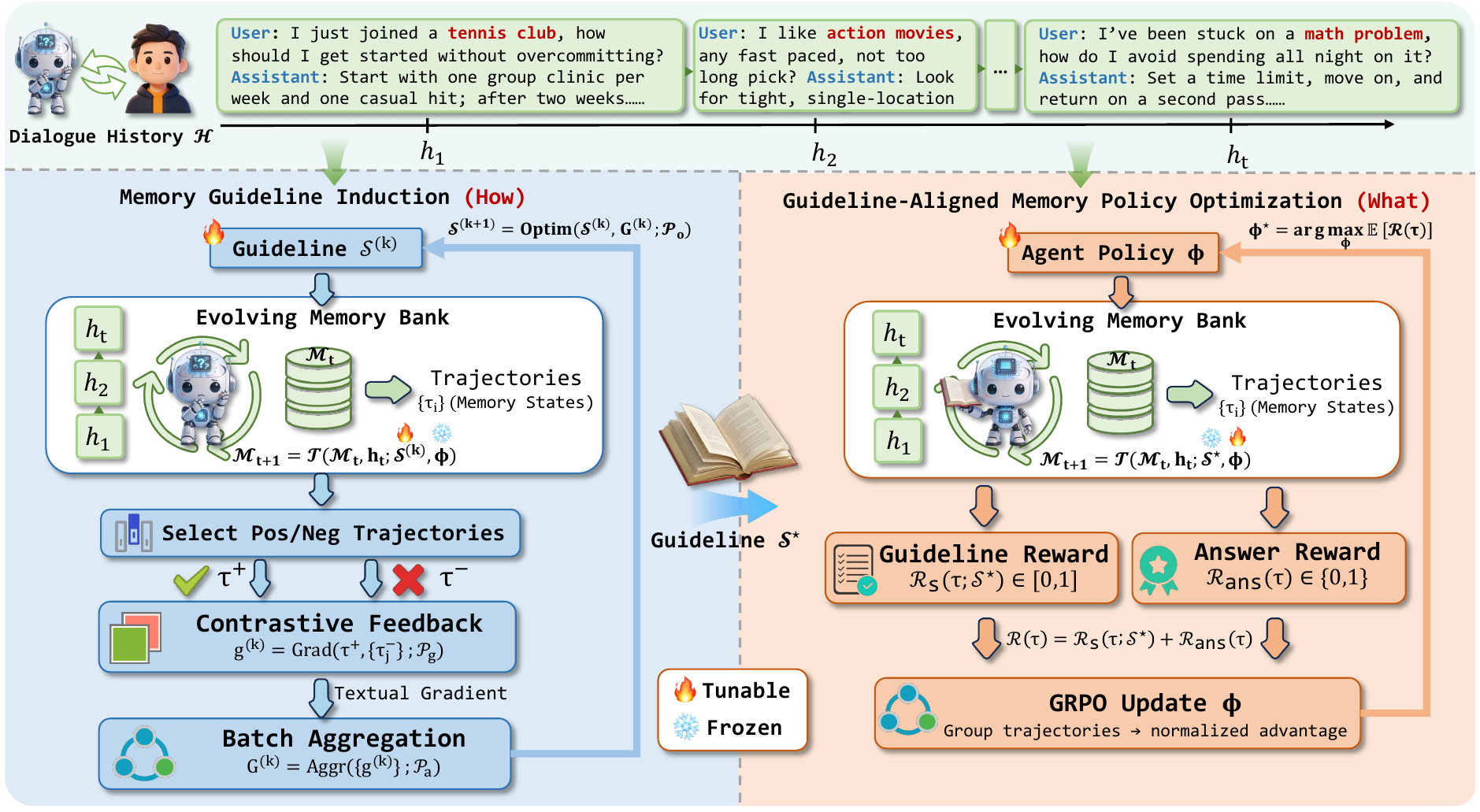}
\caption{Overview of our proposed \ourmodel. It performs two-stage optimization for evolving user memory: (1) {Memory Guideline Induction} (\moduleone) iteratively refines a natural-language guideline; (2) {Guideline-Aligned Memory Policy Optimization} (\moduletwo) fixes the induced guideline to define guideline-aligned rewards and applies multi-turn GRPO to learn what information to update in evolving memory bank.}
\label{fig:method}
\end{figure*}

We propose a two-stage framework for learning an effective memory-evolution mechanism. Instead of hand-crafting the update rule inside the evolution operator \( \mathcal{T} \), we treat the memory update instruction as an optimizable natural-language parameter and learn it from data. In the first stage, \textbf{Memory Guideline Induction}, we learn how the agent performs memory operations by inducing a high-quality textual guideline. Subsequently, we further optimize what to store in accordance with this guideline. The two stages are shown in Figure~\ref{fig:method}.

\subsection{Memory Guideline Induction}
\label{subsec:memory_guideline}

Existing implementations of memory-evolution methods typically rely on manually designed templates or prompts that prescribe how new dialogue segments should modify the user's memory. Such heuristic guidelines are brittle and lack the ability to adapt across domains, user styles, and annotation conventions. Inspired by schema-based human memory mechanisms, we instead treat the instruction prompt \( \mathcal{S} \) as a global natural-language parameter encoding a structured policy for memory operations, and aim to learn it from data. Consequently, the objective of the Memory Guideline Induction stage is therefore to induce an optimized guideline \( \mathcal{S}^\star \) that teaches the agent how to perform memory evolution correctly.

\paragraph{Contrastive feedback as textual gradient.}
Firstly, we use a training set where each example provides a dialogue history \(\mathcal{H}\) and a query \(x\). At optimization step \(k\), given the current guideline \(\mathcal{S}^{(k)}\), we first run the memory-evolution operator over the history and then perform multiple forward propagations of the agent to answer the query. This produces a set of trajectories \(\{\tau_i\}\), where each trajectory \(\tau_i\) contains the query \(x\), the intermediate memory states, and a candidate response \(y_i\). Using task supervision or environment feedback, we select at least one correct trajectory \(\tau^{+}\) and treat the remaining, partially plausible but suboptimal ones as contrastive negatives \(\{\tau_j^{-}\}\).
To obtain contrastive feedback, we apply a predefined feedback instruction \(\mathcal{P}_g\) that compares the correct trajectory \(\tau^{+}\) with the negative trajectories \(\{\tau_j^{-}\}\), highlighting the desired properties of \(\tau^{+}\) and the typical errors in the negatives. The resulting natural-language contrastive reflection serves as a \textbf{textual gradient}, guiding the iterative refinement of the guidelines:
\begin{equation}
g^{(k)} = \mathrm{Grad}\big(\tau^{+}, \{\tau_j^{-}\} ; \mathcal{P}_g\big).
\end{equation}
This textual gradient is then used to update \(\mathcal{S}^{(k)}\), guiding the agent toward more reliable and task-aligned trajectories.

\paragraph{Batch-level gradient aggregation.}
To obtain a stable and general update signal, we aggregate textual gradients across a mini-batch \(B\) of training examples. Each \(g^{(k)}\) provides a localized critique about how \(\mathcal{S}^{(k)}\) should change for a specific \((\mathcal{H}, x)\); the $\mathrm{Aggr}(\cdot)$ operator synthesizes these instance-level signals into a single, abstract update direction:
\begin{equation}
G^{(k)} = \mathrm{Aggr}\big(\{g^{(k)}\}_{(\mathcal{H},x)}; \mathcal{P}_a\big),
\end{equation}
where \(\mathrm{Aggr}(\cdot)\) can be instantiated as a summarization and abstraction procedure, guided by an aggregation prompt \(\mathcal{P}_a\), that identifies common failure patterns and consolidates them into a guideline-level modification proposal.

\paragraph{Optimization objective.}
By applying the merged textual gradient \(G^{(k)}\), the guideline \(\mathcal{S}^{(k)}\) is refined through an optimization operator that performs natural-language editing.
\begin{equation}
\mathcal{S}^{(k+1)} = \mathrm{Optim}\big(\mathcal{S}^{(k)}, G^{(k)};\mathcal{P}_o\big).
\end{equation}
Conceptually, this iterative procedure performs gradient-like steps on an underlying contrastive objective that promotes answers aligned with the positive references and penalizes confusing them with the negatives. Let \(\mathcal{R}(\cdot)\) denote a reward function; in our implementation, it simply indicates whether the output is correct. Under this view, the induced guideline \(\mathcal{S}^\star\) can be regarded as an approximate maximizer of the expected reward:
\begin{equation} 
\mathcal{S}^{\star}
= \arg\max_{\mathcal{S}}
\mathbb{E}_{(\mathcal{H}, x)}
\Big[
\mathcal{R}\big(\tau^{+}, \{\tau_j^{-}\} ; \mathcal{S}\big)
\Big].
\end{equation}
This yields a memory guideline that encodes effective principles for guiding downstream memory evolution, which are provided in Figure \ref{prompt:template_evolve_step50}.

\subsection{Guideline-Aligned Memory Policy Optimization}

Building on the induced guideline \( \mathcal{S}^\star \), the second stage focuses on optimizing \emph{what} to store in the user memory. We fix \( \mathcal{S}^\star \) and regard the parameters \( \phi \) of the evolution operator \( \mathcal{T} \) and the agent \( \mathcal{A} \) as a unified policy over memory-augmented trajectories. For each training instance \((\mathcal{H}, x)\), rolling out the system under \( \mathcal{S}^\star \) produces a trajectory \(\tau\) that interleaves memory updates
\(
\mathcal{M}_{t+1} = \mathcal{T}(\mathcal{M}_t, h_t ; \mathcal{S}^\star, \phi)
\)
and intermediate responses
\(
y_t = \mathcal{A}(x, \mathcal{M}_t),
\)
culminating in a final answer used for evaluation.

\paragraph{Guideline-aligned rewards.}
Our first signal is a \emph{Guideline-aware} reward that explicitly enforces the guideline induced by \( \mathcal{S}^\star \). For each memory-update segment in \(\tau\), we parse the model output and prompt LLM to score whether the update strictly follows the prescribed output format (e.g., required fields, tags, and structure). These signals are aggregated into a dense guideline reward
\(
\mathcal{R}_{\text{S}}(\tau; \mathcal{S}^\star) \in [0,1],
\)
which encourages \(\mathcal{T}\) to produce guideline-aligned, well-structured memory edits rather than arbitrary free-form text. 
 Second, an answer reward \( \mathcal{R}_{\text{ans}}(\tau)  \in \{0,1\}\) measures task correctness by directly comparing the final response in \(\tau\) with the reference answer (e.g., exact or judged match), yielding a simple correctness signal used to align the memory policy with downstream performance. The overall trajectory reward combines the two components as
\(
\mathcal{R}(\tau) = (1-\lambda) * \mathcal{R}_{S}(\tau; \mathcal{S}^\star) + \lambda *  \mathcal{R}_{\text{ans}}(\tau),
\)
where \( \lambda \) balances guideline fidelity and answer accuracy.

\paragraph{Policy optimization.}
 We optimize \( \phi \) using Group Relative Policy Optimization (GRPO) over groups of trajectories on multi-conversation memory evolution. For each \((\mathcal{H},x)\), GRPO samples a group of trajectories, computes group-normalized advantages from \( \mathcal{R}(\tau) \), and applies a clipped policy-gradient update. Abstractly, the learned guideline-aligned memory policy is obtained by
\begin{equation}
\phi^\star
=
\arg\max_{\phi}
\mathbb{E}_{(\mathcal{H},x)\sim\mathcal{D},\;
\tau \sim \pi_\phi(\cdot \mid \mathcal{H},x;\mathcal{S}^\star)}
\big[
\mathcal{R}(\tau) 
\big].
\end{equation}
More details can be seen in Appendix \ref{app:grpo}. 
In this way, the second stage learns a memory-evolution policy that follows the induced guideline while selectively storing information that is most beneficial for downstream interaction quality.

\section{Experiments}\label{Experiments}

\subsection{Experimental Settings}

\begin{table*}[!htbp]
\centering
\setlength{\tabcolsep}{5.5pt}
\begin{tabular}{l|cc|cc|cccc|c}
\toprule
\multirow{2}{*}{\textbf{Method}} & \multicolumn{2}{c|}{\cellcolor{red!12}\textbf{PersonaMem}} & \multicolumn{2}{c|}{\cellcolor{blue!12}\textbf{PrefEval}} & \multicolumn{4}{c|}{\cellcolor{orange!12}\textbf{PersonaBench (Noise Level)}} & \cellcolor{cvprblue!20} \\
 & \cellcolor{red!12}32K & \cellcolor{red!12}128K & \cellcolor{blue!12}Explicit & \cellcolor{blue!12}Implicit & \cellcolor{orange!12}w/o Noise & \cellcolor{orange!12}0.3 & \cellcolor{orange!12}0.5 & \cellcolor{orange!12}0.7 & \multirow{-2}{*}{\cellcolor{cvprblue!20}\textbf{Overall}} \\ 
\midrule
Long Context & 34.36 & 25.05 & 31.70 & 30.80 & 29.00 & 19.10 & 17.83 & 13.00 & 26.90 \\
RAG          & 48.67 & 38.90 & 47.80 & 32.40 & 29.09 & 28.16 & 24.31 & 23.00 & 36.68 \\
\midrule
Mem0         & 48.53 & 39.67 & 57.60 & 46.40 & 17.60 & 19.75 & 19.22 & 17.80 & 38.23 \\
A-Mem        & 48.26 & 38.22 & 62.30 & 52.80 & 30.32 & 28.56 & 25.19 & 24.45 & 42.64 \\
LightMem     & 50.72 & 39.93 & 64.20 & 54.80 & 19.08 & 18.74 & 19.65 & 17.80 & 41.21 \\
\midrule
MemAgent     & 53.58 & 43.59 & 72.30 & 63.60 & 20.05 & 19.36 & 16.51 & 17.92 & 45.00 \\
Mem-$\alpha$ & 53.37 & 42.86 & 71.90 & 62.50 & 19.92 & 17.02 & 16.43 & 15.59 & 44.19 \\
\textbf{\ourmodel(Ours)} & \textbf{57.06} & \textbf{47.24} & \textbf{81.30} & \textbf{69.90} & \textbf{32.27} & \textbf{29.89} & \textbf{25.99} & \textbf{25.09} & \textbf{52.02} \\
\bottomrule
\end{tabular}
\caption{\textbf{Overall comparison across eight evaluation settings.} We report results on PersonaMem (32K/128K) (In-Domain), PrefEval (Explicit/Implicit) (Out-of-Domain), and PersonaBench under different noise levels (Out-of-Domain). Higher is better. The best results are highlighted in \textbf{bold}.}
\label{tab:overall_eval}
\end{table*}

\textbf{Datasets and Metrics.}
We evaluate on three personalization memory benchmarks: {PersonaMem}~\cite{personamem}, {PrefEval}~\cite{prefeval}, and {PersonaBench}~\cite{tan2025personabench}. {PersonaMem} measures preference evolution over long multi-session histories at different context scales. PrefEval emphasizes explicit vs.\ implicit preference multi-choice queries (1{,}000 each) with 50 inserted turns. PersonaBench tests personalized retrieval and QA over heterogeneous, noisy user corpora. We report accuracy on PersonaMem and PrefEval, and F1 on PersonaBench. Details are reported in Appendix \ref{app:Datasets}.

\textbf{Baselines.}
We compare our approach against a diverse set of baselines.
\textbf{LongContext} directly feeds as much of the raw interaction history.
\textbf{RAG} denotes a retrieval-augmented generation setup that indexes all historical dialogue snippets in a vector store and retrieves the top-$K$ relevant segments.
To study external memory architectures, we further include three retrieval-based memory methods, \textbf{Mem0}, \textbf{A-Mem}, and \textbf{LightMem}, which maintain an external memory bank and update it.
We also compare with two reinforcement-learning-based memory agents, \textbf{MemAgent} and \textbf{MEM-$\alpha$}, which explicitly learn memory evolution actions.
All baselines are implemented on top of the same backbone and evaluated under the same data split and hyperparameter setup for a fair comparison.

\colorlet{pruneA}{orange!5} 
\colorlet{pruneB}{orange!10} 
\colorlet{pruneC}{red!12} 
\colorlet{pruneD}{red!15} 
\colorlet{pruneE}{red!20} 
\colorlet{pruneF}{red!25}
\begin{table}[t]
\centering
\small
\setlength{\tabcolsep}{1pt} 
\begin{tabular}{p{1.6cm}|M{1.4cm}M{1.4cm}|M{1.4cm}M{1.4cm}}
\toprule
\multirow{2}{*}{\textbf{Setting}} &
\multicolumn{2}{c|}{\textbf{PersonaMem}} &
\multicolumn{2}{c}{\textbf{PrefEval}} \\
& \textbf{32K} & \textbf{128K} & \textbf{Explicit} & \textbf{Implicit} \\
\midrule
\rowcolor{pruneF}\ourmodel          & \textbf{57.06} & \textbf{47.24} & \textbf{81.30} & \textbf{69.90} \\
\rowcolor{pruneE}w/o CF    & 56.44 & 46.33 & 78.30 & 68.10 \\
\rowcolor{pruneD}w/o GR       & 56.24 & 46.06 & 79.50 & 68.30 \\
\rowcolor{pruneC}w/o MGI    & 54.81 & 44.50 & 73.20 & 63.60 \\
\rowcolor{pruneB}w/o GMPO & 53.37 & 43.97 & 77.40 & 66.20 \\
\rowcolor{pruneA}w/o ALL         & 48.47 & 39.09 & 71.70 & 60.60 \\
\bottomrule
\end{tabular}
\caption{\textbf{Ablation Study.} \textbf{CF}: a contrastive feedback for textual-gradient guideline induction. \textbf{GR}: a guideline reward for enforcing the induced schema during memory updates.}
\vspace{-1em}
\label{tab:ablation}
\end{table}

\textbf{Implementation Details.}
We mainly leverage \texttt{Qwen2.5-7B-Instruct} \cite{yang2024qwen2} as the backbone LLM for all methods (\textsc{Mem-$\alpha$}  uses \texttt{Qwen3-4B}). For retrieval, we adopt \texttt{all-MiniLM-L6-v2}~\cite{wang2020minilm} and retrieve the Top-10 candidates. We construct training data by sampling 300 examples from PersonaMem. During training, we use retrieved dialogues as context to reduce computation when learning memory evolution; during inference, we feed the full dialogue history as context. Each memory-evolving round inputs a 4K-token chunk. All baselines are implemented using their publicly available codebases. For a fair comparison, MemAgent and MEM-$\alpha$ use their publicly released checkpoints, additionally training the same 300 PersonaMem training samples used by our method. All experiments are conducted on four A6000 GPUs. The same hyperparameters are shared across all retrieval-based methods. Hyperparameters are reported in Appendix~\ref{app:hyperparameter}.

\begin{figure}[ht]
\centering
\includegraphics[width=0.88\linewidth]{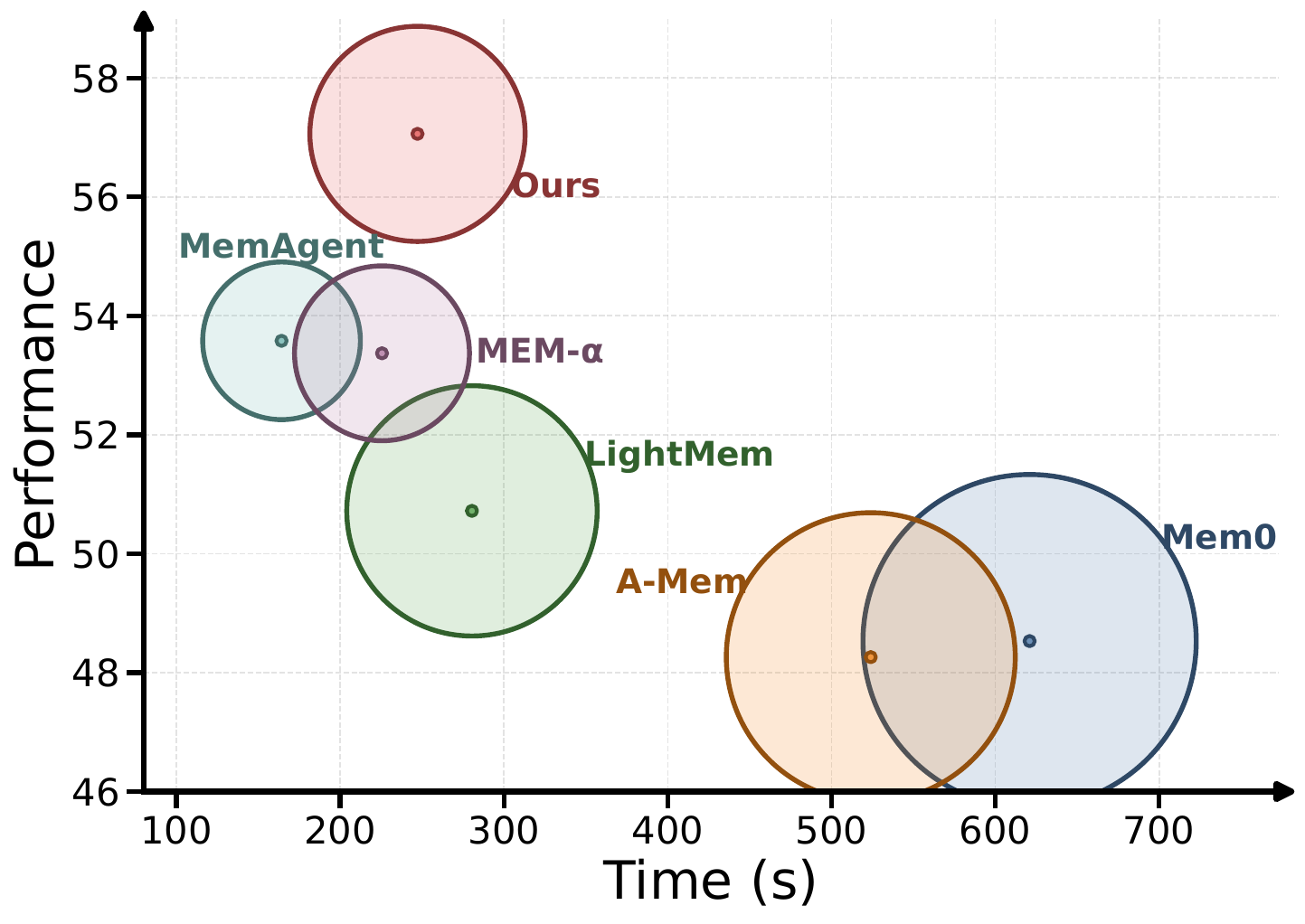}
\caption{Efficiency analysis on PersonaMem. We report the performance--time balance of memory construction/evolution over 20 dialogue histories (32K), where circle size indicates the standard deviation of runtime.}
\vspace{-1em}
\label{fig:efficiency_tradeoff}
\end{figure}

\subsection{Overall Evaluation}
\label{subsec:overall_eval}

\paragraph{Overall Comparison with Baselines.}
Table~\ref{tab:overall_eval} shows that \ourmodel achieves the best overall score across the three benchmarks, indicating that learning a memory-evolution mechanism is more effective than fixed context inclusion or manually designed update heuristics. Specifically, Long Context degrades notably under noisy histories, while \ourmodel captures user preference by filtering irrelevant content when evolving memory. 
Compared with explicit memory-bank baselines (Mem0, A-Mem, LightMem), \ourmodel delivers larger improvements on both PersonaMem and PrefEval. RL-based memory agents (MemAgent, Mem-$\alpha$) are competitive, yet they still lag behind in overall performance. This trend aligns well with our two-stage design: \moduleone induces a transferable guideline for memory evolution, while \moduletwo learns to retain preference-relevant information under the guideline, which demonstrates the effectiveness of our method for stable long-horizon personalization.
\paragraph{Generalizations Across Settings.}
Across the eight settings in Table~\ref{tab:overall_eval}, \ourmodel shows strong generality, consistently outperforming baselines on both in-domain tasks (PersonaMem, 32K$\rightarrow$128K) and out-of-domain tasks (PrefEval Explicit/Implicit; PersonaBench with increasing noise) evaluations, consistent with our two-stage design: \moduleone learns stable memory organizations, while \moduletwo retains preference-relevant information. We reported results in different categories in Appendix \ref{app:ComparisoninDifferentCategories}.

\begin{table}[t]
\centering
\small
\setlength{\tabcolsep}{5pt}
\begin{tabular}{l|c c c c}
\toprule
\textbf{Method} & \makecell{\textbf{Qwen2.5-7B}\\\textbf{Instruct}}
& \makecell{\textbf{gpt-4o}\\\textbf{-mini}}
& \makecell{\textbf{gemini-2.5}\\\textbf{-flash}}
& \textbf{GPT-5} \\
\midrule
RAG & 48.67 & 47.44 & 61.15 & 63.80 \\
A-Mem & 48.26 & 48.47 & 62.37 & 64.42 \\
\midrule
 \multicolumn{5}{l}{\cellcolor{cvprblue!7}\textit{$\blacktriangledown$ Optimized w/ Qwen2.5-7B-Instruct}} \\
\ourmodel & \textbf{53.37} & 52.56 & 64.62 & 66.67 \\
\midrule 
 \multicolumn{5}{l}{\cellcolor{cvprblue!7}\textit{$\blacktriangledown$ Optimized w/ gpt-4o-mini}} \\
\ourmodel & 52.56 & \textbf{54.19} & \textbf{64.83} & \textbf{67.28} \\
\bottomrule
\end{tabular}
\caption{Cross-LLM transferability of \moduleone optimized guidelines (without RL). We optimize the guideline with one LLM and evaluate with different LLMs.}
\vspace{-1em}
\label{tab:cross_llm_transfer_textgrad}
\end{table}

\subsection{Ablation Study}

To further investigate the impact of each designed module, we conduct ablation study on two prevalent datasets. As depicted in Table~\ref{tab:ablation}, it shows that the full model performs best on both long-context PersonaMem and distractor-heavy PrefEval, indicating that our two-stage framework is necessary for stable memory evolution. Removing \textbf{CF} or \textbf{GR} causes consistent but smaller drops (e.g., PersonaMem 32k: 57.06$\rightarrow$56.44 / 56.24; PrefEval Explicit: 81.30$\rightarrow$78.30/79.50), suggesting that contrastive textual feedback and guideline-aligned rewards both improve update reliability. In contrast, ablating either stage yields more significant degradations: removing \textbf{\moduleone} most strongly hurts preference retention and inference performance (PrefEval Explicit/Implicit: 81.30/69.90$\rightarrow$73.20/63.60), while removing \textbf{\moduletwo} more severely impacts long-horizon tracking on PersonaMem (32k/128k: 57.06/47.24$\rightarrow$53.37/43.97). Finally, \textbf{w/o ALL} collapses performance across benchmarks (PersonaMem 32k: 48.47; PrefEval Explicit: 71.70), confirming that learned guidelines plus guideline-aligned policy optimization are both critical.

\subsection{Efficiency Analysis}

Considering the use of LLMs, we further explore the efficiency of our proposed method. Figure~\ref{fig:efficiency_tradeoff} illustrates the performance-time trade-off of memory construction/evolution. Our \ourmodel achieves the best performance while remaining among the faster approaches, indicating a favorable efficiency frontier rather than a pure accuracy-at-any-cost gain. This advantage is consistent with the design of \ourmodel: instead of repeatedly invoking an LLM to separately extract and merge memory entries (e.g., A-Mem and Mem0), our approach \textbf{internalizes extraction, update, and forgetting behaviors into the model’s memory evolution process}, reducing integration overhead. In contrast, MemAgent and MEM-$\alpha$ run quickly but fall behind in performance, suggesting that their memory update mechanism cannot reliably maintain useful user information.

\subsection{Cross-LLM Transferability of Guidelines}

We also investigate the transferability of different LLMs in our method, particularly focusing on the LLMs used for optimization and evaluation. As shown in Table~\ref{tab:cross_llm_transfer_textgrad}, we select various mainstream LLMs to evaluate whether \textbf{optimized guidelines} transfer across different backbone LLMs. The results show that, across all four LLMs, both \moduleone variants consistently outperform the baselines (RAG and A-Mem), indicating that the learned guideline captures model-agnostic memory-update principles rather than overfitting to a specific LLM. Notably, optimizing with \texttt{gpt-4o-mini} generalizes strongly and achieves the best numbers on three backbones, including \texttt{GPT-5} and \texttt{gemini-2.5-flash}. Overall, these results support that \moduleone produces a guideline that is portable across LLMs, making it practical to optimize once and deploy under different backbone choices.

\begin{figure}[t]
\centering
\includegraphics[width=0.9\linewidth]{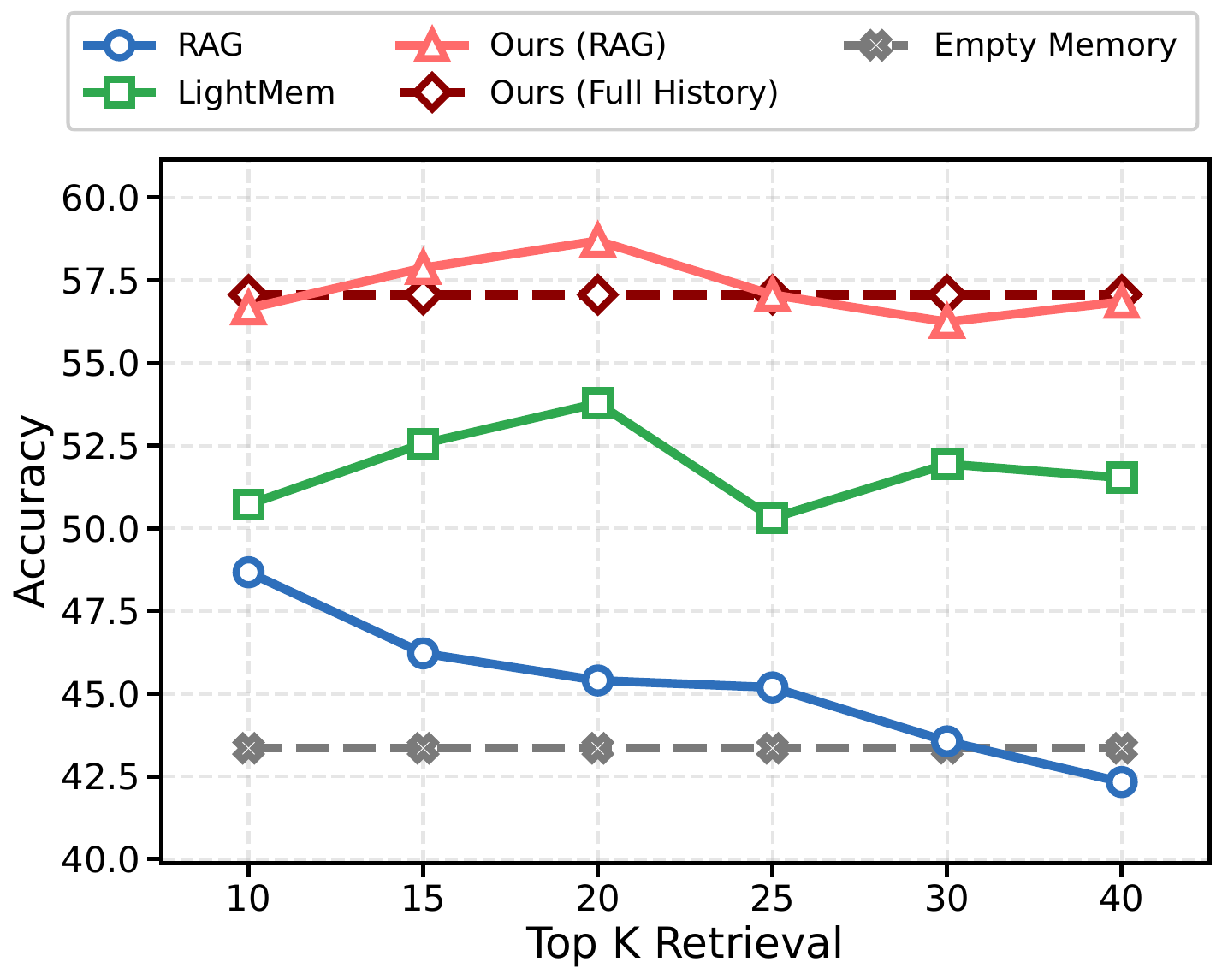}
\caption{Retrieval Top-$K$ on PersonaMem (32K).}
\vspace{-1em}
\label{fig:topk}
\end{figure}

\subsection{Comparison on Different Retrieval}
In Figure~\ref{fig:topk}, we examine how Top-$K$ retrieval affects different methods on the PersonaMem dataset.
Across all Top-$K$, both inference modes of our approach (performing memory evolution on retrieved context, \textbf{Ours (RAG)}, or on full history, \textbf{Ours (Full History)}) remain consistently strong and clearly outperform the baselines.
Notably, \textbf{Ours (RAG)} peaks around $K{=}20$ and even surpasses the full-history variant, which suggests that retrieval can be beneficial when it filters irrelevant context and reduces noise for better memory evolution. In contrast, vanilla RAG degrades as $K$ increases and even drop below Empty Memory, indicating that simply adding more retrieved content may introduce distractors that hurt downstream decisions. Overall, these results show that retrieval is not sufficient on its own; it achieves its best effect when coupled with \ourmodel to transform retrieved evidence into coherent memory.

\subsection{Impact of Per-Round Token Budget on Memory Evolution}
\begin{figure}[t]
\centering
\includegraphics[width=0.98\linewidth]{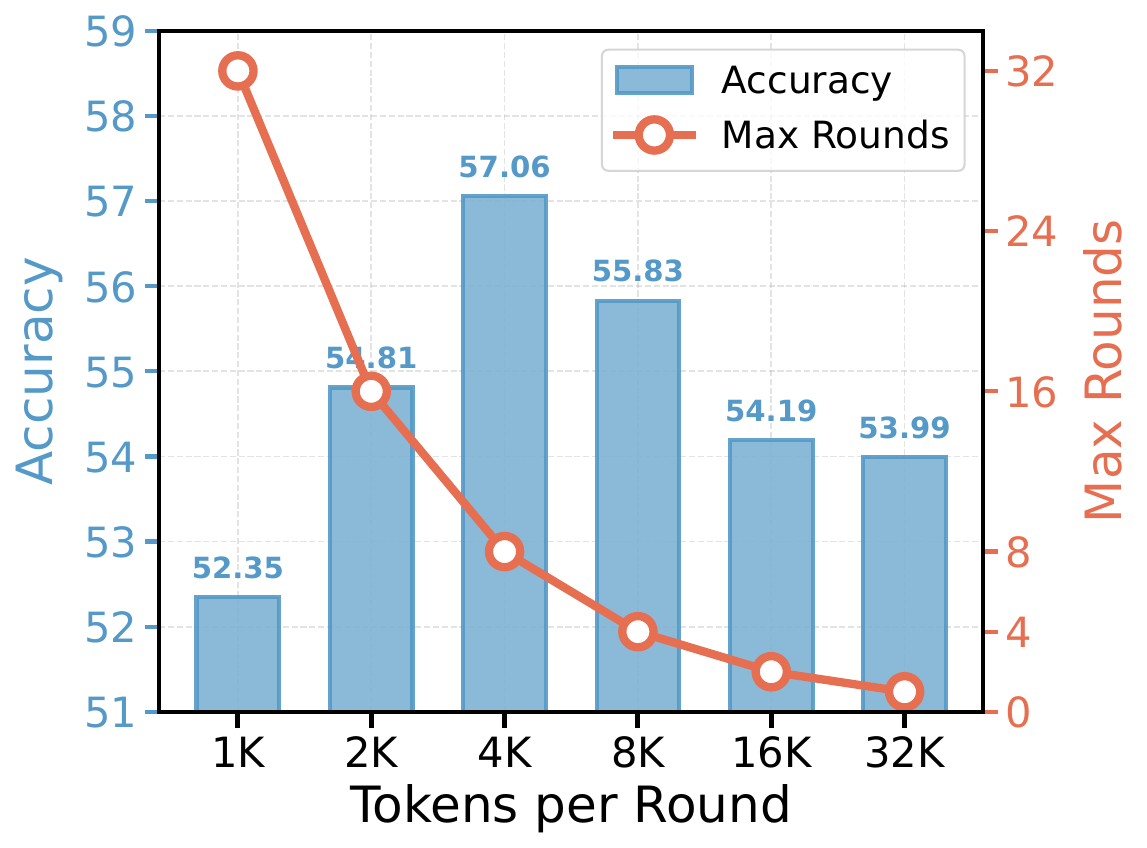}
\caption{Effect of tokens per evolve round on PersonaMem (32K).}
\vspace{-1em}
\label{fig:tokens_rounds}
\end{figure}

Since the per-round token budget directly determines inference cost in real-world deployment, we study its impact on memory evolution in Figure~\ref{fig:tokens_rounds}.
When token budget is too small (e.g., 1K--2K), system must split history into many rounds, and repeated evolve operations can accumulate errors and trigger uncontrolled forgetting, which ultimately hurts accuracy. In contrast, increasing budget initially tends to improve performance by reducing the required number of rounds and stabilizing the update dynamics. However, excessively large budgets (8K--32K) can make each evolution step more challenging, as the resulting context becomes more complex, thereby increasing the difficulty of processing information within a single pass. Overall, the results suggest a clear trade-off: effective memory evolution requires a moderate per-round token budget that avoids both excessive update frequency and overly complex single-step contexts.

\subsection{Effect of Guideline Quality}
\begin{figure}[t]
\centering
\includegraphics[width=0.89\linewidth]{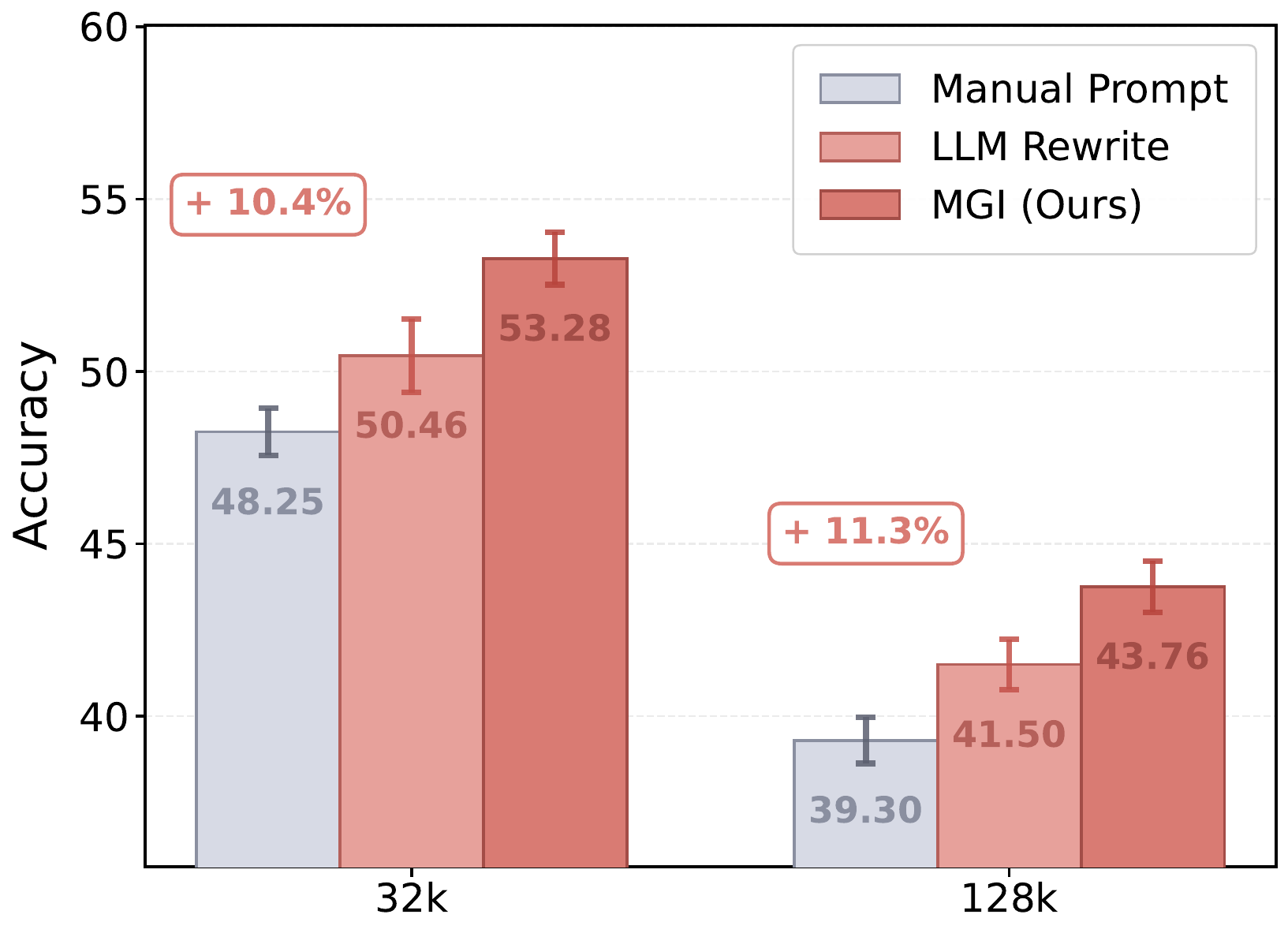}
\caption{Impact of prompt quality on PersonaMem, averaged over three random runs with different seeds; error bars indicate standard deviation.}
\vspace{-1em}
\label{fig:prompt}
\end{figure}

Finally, to gain a more intuitive understanding of the impact of guideline quality, we compare guidelines of different quality levels.
As shown in Figure~\ref{fig:prompt}, improving the quality of the memory-update guideline consistently strengthens downstream performance. Starting from a manually written prompt, an LLM rewrite yields a moderate gain, suggesting that surface-level prompt refinement helps but remains limited. Note that, the guideline induced by \moduleone achieves the best results on both settings, reaching 53.28 on 32K and 43.76 on 128K, which corresponds to relative improvements of +10.4\% and +11.3\% over the manual prompt, respectively. The error bars across three seeds indicate that these gains are stable rather than driven by randomness.

\section{Conclusion}

Inspired by memory schema theory that highlights \emph{prefrontal regions} and \emph{hippocampus regions}, we present \ourmodel, a two-stage optimization framework that decouples \textbf{how to organize memory} from \textbf{what to store}. Specifically, \ourmodel first induces a transferable, schema-consistent guideline for memory evolution, and then optimizes a guideline-aligned memory policy to decide what to retain, update, or forget across multi-session interactions. Extensive experiments on three personalization memory benchmarks show that \ourmodel consistently outperforms strong retrieval-based memory bank, and RL-based memory-agent baselines, while remaining robust under longer histories and noisier evidence. Overall, the results support that coupling an explicit evolution guideline with policy optimization yields a practical improvement in \textbf{efficiency}, \textbf{robustness}, and \textbf{transferability} for evolving user memory in conversational agents.

\section*{Acknowledgements}
This work was supported in part by the grants from National Science and Technology Major Project (No. 2023ZD0121104), National Natural Science Foundation of China (No. U22B2059), the Anhui Natural Science Foundation (No. 2508085ZD006), National Natural Science Foundation of China (No.62502404), Hong Kong Research Grants Council (Research Impact Fund No.R1015-23, Collaborative Research Fund No.C1043-24GF, General Research Fund No. 11218325), Institute of Digital Medicine of City University of Hong Kong (No.9229503), Huawei (Huawei Innovation Research Program), Tencent (Tencent Rhino-Bird Focused Research Program, Tencent University Cooperation Project), Didi (CCF-Didi Gaia Scholars Research Fund), Kuaishou (CCF-Kuaishou Large Model Explorer Fund No. 2025008, Kuaishou University Cooperation Project), and Bytedance.

\section*{Limitations}
Overall, our method is effective for improving long-horizon personalization memory by learning for more structured and consistent memory evolution. However, the second-stage optimization relies on an LLM-based scorer to provide guideline-aligned process rewards, which makes performance sensitive to scorer reliability. Moreover, our method requires careful tuning of the per-round token budget and the number of evolution rounds; when long histories are split into many rounds, small update errors can compound over time and lead to unintended forgetting or over-generalized memory entries. Finally, our current design treats memory evolution as a single-objective policy under a fixed guideline; extending it to explicitly balance multiple competing objectives (e.g., stability vs.\ plasticity, informativeness vs.\ brevity) remains non-trivial and may require additional control mechanisms.

\section*{Ethical considerations}
Our method is a general memory-evolution framework intended to support personalized agents, and it primarily improves \emph{how} existing memories are organized and optimized rather than expanding the scope of the LLM system's access. The primary ethical risk arises from misuse rather than from the method itself: if deployed without appropriate safeguards, persistent memory could be used to over-collect user information or to enable unwanted profiling. In responsible deployments, memory should follow data-minimization principles, avoid storing sensitive identifiers, and provide clear user controls for inspection, correction, and deletion; additionally, retention policies and access control should be enforced at the system level to ensure the system remains aligned with privacy expectations as application contexts evolve.

\bibliography{custom}

\newpage
\appendix

\section{GRPO for Memory Evolution}
\label{app:grpo}

For completeness, we summarize the GRPO objective for the multi-turn memory evolution process. Following MemAgent~\citep{yu2025memagent}, we optimize memory evolution over groups of trajectories. For a given input \((\mathcal{H}, x)\), the current policy \( \pi_\phi \) generates a group of \(G\) trajectories \(\{\tau_i\}_{i=1}^G\), with corresponding rewards \(\{R_i\}_{i=1}^G\). In the multi-conversation setting, each trajectory \(\tau_i\) is further decomposed into \(n_i\) conversations,
\[
\tau_i = \{\tau_{i,1}, \tau_{i,2}, \dots, \tau_{i,n_i}\},
\]
where \(\tau_{i,j}\) denotes the token sequence of the \(j\)-th conversation. GRPO normalizes rewards within each group and defines a group-relative advantage:
\begin{equation}
\widehat{A}_i
=
\frac{R_i - \mathrm{mean}(\{R_j\}_{j=1}^G)}{\mathrm{std}(\{R_j\}_{j=1}^G)}.
\end{equation}
This advantage is then assigned to all token-level actions in \(\tau_i\), including both memory-update tokens and answer tokens across all conversations. Let \(r_{i,j,t}(\phi)\) denote the importance-sampling ratio between the current policy and a frozen reference policy \(\pi_{\text{ref}}\) at token step \(t\) in the \(j\)-th conversation of trajectory \(\tau_i\). The multi-conversation GRPO objective is written as
\begin{equation}
\begin{aligned}
&J_{\mathrm{GRPO}}(\phi)
=
\mathbb{E}\Bigg[
\frac{1}{G}
\sum_{i=1}^G
\frac{1}{\sum_{j=1}^{n_i} |\tau_{i,j}|}
\sum_{j=1}^{n_i}
\sum_{t \in \tau_{i,j}}
\\[-2mm]
&\min\!\Big(
r_{i,j,t}(\phi)\widehat{A}_i,
\mathrm{clip}\!\big(r_{i,j,t}(\phi),1-\epsilon,1+\epsilon\big)\widehat{A}_i
\Big)
\\
&\qquad
-\beta\,\mathrm{KL}\!\left(\pi_\phi\,\|\,\pi_{\text{ref}}\right)
\Bigg].
\end{aligned}
\end{equation}

\section{Hyperparameter Settings} \label{app:hyperparameter}

We summarize the hyperparameter settings for training and inference in Table~\ref{tab:hyperparameters}.

\begin{table}[t]
\centering
\small
\setlength{\tabcolsep}{4pt}
\renewcommand{\arraystretch}{1.05}
\begin{tabular}{l l}
\toprule
\textbf{Phase} & \textbf{Hyperparameters} \\
\midrule
\textbf{Training (Stage 1)} & Context = RAG \\
& Round size = 512 \\
& Optimization steps = $\{10, 30, 50, 70\}$ \\
& Temperature = 1.0 \\
& Top-$p$ = 1.0 \\
& Max output tokens = 2048 \\
\midrule
\textbf{Training (Stage 2)} & Context = RAG \\
& Batch size = 4 \\
& Round size = 512 \\
& Learning rate = $1 \times 10^{-6}$ \\
& Temperature = 1.0 \\
& Top-$p$ = 1.0 \\
& Rollout batch size = 8 \\
& Rollout $n$ = \{2,4,8\} \\
& Epochs = 5 \\
& Max output tokens = 2048 \\
\midrule
\textbf{Inference} & Context = Full history \\
& Round size = $\{1\textsc{k}, 2\textsc{k}, 4\textsc{k}, 8\textsc{k}, 16\textsc{k}\}$ \\
& Serving = vLLM \\
& Max output tokens = 2048 \\
& Temperature = 0.0 \\
\bottomrule
\end{tabular}
\caption{Hyperparameter settings used for training and inference.}
\label{tab:hyperparameters}
\end{table}

\section{Datasets} 
\label{app:Datasets}

\subsection{PersonaMem Dataset}
\label{app:personamem-dataset}

PersonaMem \cite{personamem} is a large-scale benchmark for evaluating long-term personalization in conversational LLMs. It contains interaction histories for \textbf{20 simulated personas}, each designed with rich static attributes (e.g., demographics and occupation) and dynamic traits and preferences that evolve over time across \textbf{15 diverse real-world task domains} such as food recommendation, travel planning, and therapy consultation. For every persona, multi-session conversations are constructed in which the user engages with a chatbot over \textbf{7 types of in-situ queries} that probe different personalization capabilities (e.g., recalling user facts, tracking preference evolution, and providing preference-aligned suggestions). Each session consists of 15-30 user–assistant turns, and histories are instantiated at three context scales by concatenating 10, 20, or 60 sessions, yielding approximate context lengths of 32k, 128k, and 1M tokens, respectively. At evaluation time, models must select appropriate responses to user queries conditioned on the interaction history, thereby testing their ability to evolve over dynamic user profiles. The main statistics of PersonaMem are summarized in Table~\ref{tab:personamem-stats}.

\begin{table}[t]
\setlength{\tabcolsep}{4pt}
  \centering
  \begin{tabular}{lccc}
    \toprule
    \textbf{Statistic} & \multicolumn{3}{c}{\textbf{PersonaMem}} \\
    \midrule
    Tokens per history        & \textasciitilde 32k   & \textasciitilde 128k  & \textasciitilde 1M     \\
    \# QA pairs               & 589   & 2727  & 2674   \\
    \# Sessions per history   & 10    & 20    & 60     \\
    Avg.\ \# utterances       & 167.1 & 758.3 & 3607.9 \\
    \bottomrule
  \end{tabular}
  \caption{Statistics of the PersonaMem dataset at different context lengths. Token counts denote the approximate total context length per interaction history; utterance counts are averaged over histories.}
  \label{tab:personamem-stats}
\end{table}

\begin{table}[t]
\centering
\begin{tabular}{l r}
\toprule
\textbf{Statistic} & \textbf{Value} \\
\midrule
Explicit queries & 1{,}000 \\
Implicit queries & 1{,}000 \\
Maximum inserted conversations & 24 \\
Maximum inserted turns & 326 \\
Avg.\ turns / conversation & 13.58 \\
Total tokens & 108{,}102 \\
Avg.\ tokens / conversation & 4{,}504.25 \\
\bottomrule
\end{tabular}
\caption{Dataset statistics for PrefEval multiple-choice classification.}
\label{tab:prefeval_stats}
\end{table}

\begin{table}[t]
\centering
\small
\begin{tabular}{lrrrrr}
\toprule
\textbf{User} & \textbf{Queries} & \textbf{Corpus} & \textbf{Conv.} & \textbf{AI} & \textbf{E-com.} \\
\midrule
1 & 48 & 110 & 84 & 23 & 3 \\
2 & 43 & 90  & 78 & 8  & 4 \\
3 & 42 & 64  & 51 & 12 & 1 \\
4 & 46 & 85  & 71 & 14 & 0 \\
5 & 44 & 84  & 59 & 21 & 4 \\
6 & 40 & 94  & 79 & 14 & 1 \\
\midrule
\textbf{Sum} & 263 & 527 & 422 & 92 & 13 \\
\bottomrule
\end{tabular}
\caption{Statistics of the PersonaBench subset across six users. \textit{Corpus} is the sum of \textit{Conv.}, \textit{AI}, and \textit{E-com.}.}
\label{tab:personabench-stats}
\end{table}

\subsection{PrefEval Dataset}
PrefEval is a long-context, multi-session benchmark for evaluating whether LLMs can infer, retrieve, and act on user preferences in realistic conversational settings, with an emphasis on four aspects: preference inference, long-context retrieval, preference following, and personalization proactiveness. The dataset comprises 1{,}000 unique preference--query pairs, and spans 20 everyday topics grouped into seven domains: \textit{Entertainment} (Shows, Music \& Books, Sports, Games), \textit{Travel} (Activities, Restaurant, Hotel, Transport), \textit{Lifestyle} (Dietary, Beauty, Fitness, Health), \textit{Shopping} (Home, Fashion, Motors, Technology), \textit{Education} (Resources, Learn Styles), \textit{Professional Ownership}, and \textit{Professional Work Style}. PrefEval supports two evaluation formats: a free-form generation setting and a 4-way multiple-choice classification setting in which exactly one option is consistent with the stated preference. To stress long-range personalization, the benchmark inserts unrelated multi-session dialogue turns between the preference revelation and the final query. In our experiments, we use 1{,}000 explicit and 1{,}000 implicit instances under the multiple-choice classification setting, and insert 50 intervening turns as distractor context; summary statistics of our subset are reported in Table~\ref{tab:prefeval_stats}.

\subsection{PersonaBench Dataset}

PersonaBench \cite{tan2025personabench} is a benchmark designed to evaluate personalized retrieval and question answering grounded in user-specific context. For each user, it provides a heterogeneous personal corpus comprising (i) conversations with friends (\textit{Conv.}), (ii) dialogues with AI assistants (\textit{AI}), and (iii) e-commerce purchase histories (\textit{E-com.}). The evaluation queries are typically short and underspecified, requiring models to resolve implicit intent by grounding responses in evidence distributed across the user’s historical interactions and behaviors. This setting tests a model’s ability to align with diverse, user-dependent semantics under realistic contextual ambiguity. Table~\ref{tab:personabench-stats} summarizes the per-user query counts and corpus statistics for the six-user subset used in our experiments.

\section{Comparison in Different Categories} \label{app:ComparisoninDifferentCategories}

\paragraph{Comparison in Different Categories of PersonaMem.}

\begin{table*}[ht]
\centering
\small
\setlength{\tabcolsep}{3pt}
\begin{tabular}{p{2.2cm}|M{1.4cm}M{1.4cm}M{1.4cm}M{1.4cm}M{1.4cm}M{1.4cm}M{1.4cm}|M{1.4cm}}
\toprule
\textbf{Method}  &
\textbf{Recall facts} &
\textbf{Suggest ideas} &
\textbf{Latest prefs} &
\textbf{Prefs evolve} &
\textbf{Update reasons} &
\textbf{Aligned recs} &
\textbf{New Scenarios} &
\textbf{Overall} \\
\midrule
\multicolumn{9}{c}{\cellcolor{cvprblue!12}\textit{\textbf{$\blacktriangledown$ 32K memory corpus data}}} \\
\midrule
Long Context & 28.93 & 11.39 & - & 44.07 & 61.25 & 35.56 & 15.22 & 34.36 \\
RAG          & 42.98 & 15.19 & - & 59.32 & 80.00 & 51.11 & 36.96 & 48.67 \\
Mem0         & 47.93 & \textbf{19.41} & - & 46.61 & 79.58 & 57.04 & 42.75 & 48.53 \\
A-Mem        & 47.11 & 10.13 & - & 61.86 & 80.00 & 44.44 & 30.43 & 48.26 \\
LightMem     & 52.07 & 10.13 & - & 65.25 & 77.50 & 51.11 & 32.61 & 50.72 \\
MemAgent     & 54.55 &  8.86 & - & 65.25 & 76.25 & 57.78 & 54.35 & 53.58 \\
Mem-$\alpha$ & 57.02 &  7.59 & - & 64.41 & 72.50 & 60.00 & 54.35 & 53.37 \\
 \ourmodel(Ours) & \textbf{59.50} &  8.86 & - & \textbf{68.64} & \textbf{81.25} & \textbf{62.22} & \textbf{56.52} & \textbf{57.06} \\
\midrule
\multicolumn{9}{c}{\cellcolor{blue!10}\textit{\textbf{$\blacktriangledown$ 128K memory corpus data}}} \\
\midrule
Long Context & 18.12 & 23.36 & 20.62 & 38.62 & 39.77 & 21.70 & 16.99 & 25.05 \\
RAG          & \textbf{54.37} & \textbf{19.67} & 36.45 & 52.69 & 58.71 & 41.94 & 29.61 & 38.90 \\
Mem0         & 56.25 & 20.49 & 37.77 & 55.09 & 57.20 & 41.64 & 29.13 & 39.67 \\
A-Mem        & 36.88 & 16.19 & 38.73 & 59.88 & 62.50 & 35.19 & 28.16 & 38.22 \\
LightMem     & 40.00 & 17.42 & 42.33 & 61.08 & \textbf{64.02} & 35.48 & 25.73 & 39.93 \\
MemAgent     & 50.62 & 10.86 & 49.40 & 61.68 & 59.47 & 47.80 & 35.44 & 43.59 \\
Mem-$\alpha$ & 48.75 & 10.45 & 50.12 & 59.28 & 54.55 & 47.21 & 36.89 & 42.86 \\
 \ourmodel(Ours) & 52.50 & 11.68 & \textbf{54.44} & \textbf{66.47} & \textbf{64.02} & \textbf{51.61} & \textbf{38.35} & \textbf{47.24} \\
\bottomrule
\end{tabular}
\caption{Category-wise accuracy (\%) on PersonaMem under 32K and 128K interaction histories. “--” indicates the category is not available in the dataset.}
\label{tab:personamem_comparison}
\end{table*}

\begin{table*}[!htbp]
\centering
\small
\setlength{\tabcolsep}{3pt}
\begin{tabular}{p{2.2cm}|M{1.4cm}M{1.4cm}M{1.4cm}M{1.4cm}M{1.4cm}M{1.4cm}M{1.4cm}|M{1.4cm}}
\toprule
\textbf{Method}  &
\textbf{Travel} &
\textbf{Entertain} &
\textbf{Lifestyle} &
\textbf{Shop} &
\textbf{Education} &
\textbf{Professional} &
\textbf{Pet} &
\textbf{Overall} \\
\midrule
\multicolumn{9}{c}{\cellcolor{cvprblue!12}\textit{\textbf{$\blacktriangledown$ Explicit Memory}}} \\
\midrule
Long Context & 31.78 & 30.77 & 32.70 & 30.00 & 32.94 & 27.78 & 39.53 & 31.70 \\
RAG         & 45.33 & 52.04 & 47.87 & 45.79 & 42.35 & 58.33 & 48.84 & 47.80 \\
Mem0        & 58.41 & 61.99 & 57.35 & 53.68 & 54.12 & 69.44 & 46.51 & 57.60 \\
A-Mem       & 62.15 & 69.23 & 60.19 & 61.05 & 54.12 & 66.67 & 55.81 & 62.30 \\
LightMem    & 65.42 & 68.33 & 65.40 & 62.11 & 56.47 & 66.67 & 53.49 & 64.20 \\
MemAgent    & 71.03 & \textbf{77.83} & 70.62 & 71.05 & 62.35 & \textbf{83.33} & \textbf{74.42} & 72.30 \\
Mem-$\alpha$& 78.50 & 73.76 & 67.30 & 70.00 & 65.88 & 77.78 & 67.44 & 71.90 \\
\ourmodel(Ours) & \textbf{82.24} & 76.92 & \textbf{84.83} & \textbf{82.11} & \textbf{85.88} & \textbf{83.33} & 67.44 & \textbf{81.30} \\
\midrule
\multicolumn{9}{c}{\cellcolor{blue!10}\textit{\textbf{$\blacktriangledown$ Implicit Memory}}} \\
\midrule
Long Context & 30.84 & 27.60 & 34.12 & 29.47 & 34.12 & 25.00 & 34.88 & 30.80 \\
RAG         & 26.17 & 41.18 & 33.65 & 29.47 & 30.59 & 13.89 & 44.19 & 32.40 \\
Mem0        & 43.93 & 51.13 & 48.34 & 42.11 & 44.71 & 30.56 & 60.47 & 46.40 \\
A-Mem       & 51.40 & 55.20 & 55.45 & 47.37 & 54.12 & 50.00 & 58.14 & 52.80 \\
LightMem    & 50.47 & 60.63 & 58.77 & 51.58 & 47.06 & 44.44 & 65.12 & 54.80 \\
MemAgent    & 59.35 & 66.52 & 65.88 & 63.68 & 56.47 & 52.78 & \textbf{81.40} & 63.60 \\
Mem-$\alpha$& 61.68 & 66.52 & 62.09 & 61.05 & 60.00 & 52.78 & 67.44 & 62.50 \\
 \ourmodel(Ours) & \textbf{64.02} & \textbf{69.23} & \textbf{73.93} & \textbf{72.63} & \textbf{70.59} & \textbf{63.89} & 74.42 & \textbf{69.90} \\
\bottomrule
\end{tabular}
\caption{Domain-wise accuracy (\%) on PrefEval multiple-choice classification under Explicit vs.\ Implicit preference.}
\label{tab:prefeval_category}
\end{table*}

\begin{table*}[!htbp]
\centering
\small
\begin{tabular}{l|c c c c c|c c c c c}
\toprule
\textbf{Method} &
\makecell{\textbf{Basic}\\\textbf{Info}} &
\makecell{\textbf{Pref.}\\\textbf{(Easy)}} &
\makecell{\textbf{Pref.}\\\textbf{(Hard)}} &
\textbf{Social} &
\textbf{Overall} &
\makecell{\textbf{Basic}\\\textbf{Info}} &
\makecell{\textbf{Pref.}\\\textbf{(Easy)}} &
\makecell{\textbf{Pref.}\\\textbf{(Hard)}} &
\textbf{Social} &
\textbf{Overall} \\
\midrule
\multicolumn{1}{c|}{} &
\multicolumn{5}{c|}{\cellcolor{cvprblue!12}\textit{\textbf{$\blacktriangledown$ Without Noise Memory}}} &
\multicolumn{5}{c}{\cellcolor{blue!10}\textit{\textbf{$\blacktriangledown$ With 0.3 Noise Memory}}} \\
\midrule
Long Context  & \textbf{29.23} & 34.41 & 24.51 & 29.36 & 29.00 & 21.15 & 22.24 & 18.78 & 13.55 & 19.10 \\
RAG           & 24.32 & 34.06 & 32.77 & 33.68 & 29.09 & 26.14 & 39.29 & 28.63 & 26.53 & 28.16 \\
Mem0          & 19.23 & 19.09 & 18.80 & 12.58 & 17.60 & 17.41 & 29.08 & 21.87 & 18.39 & 19.75 \\
A-Mem         & 25.69 & 34.04 & 34.46 & 34.90 & 30.32 & 27.00 & \textbf{39.65} & 26.94 & \textbf{27.61} & 28.56 \\
LightMem      & 18.92 & 20.41 & 21.40 & 16.96 & 19.08 & 21.25 & 22.60 & 18.52 & 11.82 & 18.74 \\
MemAgent      & 14.32 & 31.09 & 26.58 & 21.48 & 20.05 & 16.30 & 32.36 & 18.76 & 19.77 & 19.36 \\
Mem-$\alpha$  & 14.91 & 25.92 & 30.03 & 19.55 & 19.92 & 11.87 & 27.71 & 26.00 & 15.52 & 17.02 \\
\ourmodel     & 26.74 & \textbf{34.61} & \textbf{37.02} & \textbf{38.95} & \textbf{32.27} & \textbf{29.81} & 35.12 & \textbf{29.92} & 27.48 & \textbf{29.89} \\
\midrule
\multicolumn{1}{c|}{} &
\multicolumn{5}{c|}{\cellcolor{cvprblue!12}\textit{\textbf{$\blacktriangledown$ With 0.5 Noise Memory}}} &
\multicolumn{5}{c}{\cellcolor{blue!10}\textit{\textbf{$\blacktriangledown$ With 0.7 Noise Memory}}} \\
\midrule
Long Context  & 18.65 & 16.52 & 17.90 & 16.70 & 17.83 & 11.25 & 22.26 & 18.62 & 7.75  & 13.00 \\
RAG           & 22.45 & 27.49 & 22.58 & 27.95 & 24.31 & 18.38 & 31.83 & 25.85 & 26.05 & 23.00 \\
Mem0          & 18.33 & 24.31 & 23.79 & 15.04 & 19.22 & 15.07 & 21.38 & 21.58 & 18.76 & 17.80 \\
A-Mem         & 22.28 & 23.64 & \textbf{28.12} & \textbf{29.73} & 25.19 & 18.42 & 29.96 & \textbf{28.11} & 31.42 & 24.45 \\
LightMem      & 22.85 & 20.47 & 17.09 & 14.59 & 19.65 & 20.30 & 19.80 & 17.35 & 12.00 & 17.80 \\
MemAgent      & 13.40 & 16.73 & 21.13 & 19.27 & 16.51 & 13.76 & 26.45 & 23.01 & 18.44 & 17.92 \\
Mem-$\alpha$  & 11.20 & 17.31 & 22.36 & 22.28 & 16.43 & 12.42 & 22.66 & 18.55 & 16.42 & 15.59 \\
\ourmodel     & \textbf{23.82} & \textbf{33.01} & 23.19 & 29.21 & \textbf{25.99}& \textbf{20.62} & \textbf{36.91} & 27.09 & \textbf{27.00} & \textbf{25.09} \\
\bottomrule
\end{tabular}
\caption{Category-wise macro F1 (\%) on PersonaBench for the six-user subset under different noise rates injected into the memory bank.}
\label{tab:personabench_category}
\end{table*}

Table~\ref{tab:personamem_comparison} reports category-wise results on PersonaMem under 32K and 128K interaction histories. Across both scales, \ourmodel achieves the best overall performance (57.06 at 32K; 47.24 at 128K), and the gains are concentrated on memory-dependent personalization abilities. On 32K histories, \ourmodel leads in  Recall facts (59.50), Prefs evolve (68.64), Update reasons (81.25), Aligned recs (62.22), and New Scenarios (56.52), which jointly drives a clear margin over the strongest baselines (e.g., 57.06 vs.\ 53.58 for MemAgent). When scaling to 128K, \textit{Long Context} degrades sharply overall (25.05), while memory-based methods remain substantially stronger; within them, \ourmodel stays best-performing and ranks first on Latest prefs (54.44), Prefs evolve (66.47), Aligned recs (51.61), and New Scenarios (38.35). In contrast, Suggest ideas is not a strength for \ourmodel (8.86/11.68), where methods that do not emphasize memory evolution (e.g., Mem0 or Long Context) are higher, indicating that our improvements primarily come from better tracking and applying evolving persona preferences rather than open-ended QA. Overall, the category-wise gains suggest that explicitly \emph{structuring} memory operations and then \emph{selecting} what to keep is most effective for preference-heavy queries, and the advantage becomes more pronounced as the interaction history grows longer.

\paragraph{Comparison in Different Categories of PrefEval.}

Table~\ref{tab:prefeval_category} breaks down PrefEval performance by domain under \textbf{Explicit} and \textbf{Implicit} preference settings. Under Explicit Memory, \ourmodel attains the best overall accuracy (81.30) and shows consistently strong gains on preference-heavy domains, ranking first on Travel (82.24), Lifestyle (84.83), Shop (82.11), and Education (85.88), while remaining competitive on \textbf{Entertain} (76.92) and Professional (83.33). A notable exception is Pet, where \ourmodel (67.44) trails MemAgent (74.42), suggesting that not all topics benefit equally from the same memory update behavior. Under Implicit Memory, the task becomes more challenging for all methods, yet \ourmodel again leads overall (69.90) and improves most clearly on domains that require inferring latent preferences from context, including Lifestyle (73.93), Shop (72.63), Education (70.59), and Professional (63.89). Compared with memory-bank baselines (Mem0/A-Mem/LightMem), the advantage of \ourmodel is broad across domains in both settings, indicating that it better resists long-range distractors and preserves preference-relevant signals. Overall, these domain-wise results reflect PrefEval’s construction: the inserted unrelated turns make long-range preference retrieval and faithful preference following the main bottlenecks, and \ourmodel improves most on domains where precise preference identification and consistent application are essential.

\paragraph{Comparison in Different Categories of PersonaBench.}

Table~\ref{tab:personabench_category} reports category-wise macro F1 on PersonaBench under increasing noise in the memory bank, where queries are short and underspecified, and evidence is distributed across heterogeneous user corpora (Table~\ref{tab:personabench-stats}). Without noise, \ourmodel achieves the best overall score (32.27) and is particularly strong on preference- and interaction-driven categories, leading on Pref.\ (Hard) (37.02) and Social (38.95), while remaining competitive on Pref.\ (Easy) (34.61); in contrast, Basic Info favors direct long-context inclusion (Long Context: 29.23 vs.\ \ourmodel: 26.74), suggesting that simple factual lookup is less dependent on selective memory evolution. As noise increases, all methods degrade, but \ourmodel remains the best overall method at every noise level (29.89 at 0.3; 25.99 at 0.5; 25.09 at 0.7), indicating stronger robustness to irrelevant or misleading memory entries. The category breakdown further shows that \ourmodel maintains clear advantages on Pref.\ (Easy) under heavier noise (36.91 at 0.7), while the performance gap on Pref.\ (Hard) and Social narrows against strong baselines (e.g., A-Mem), reflecting that fine-grained preference grounding and social inference are the most sensitive to noisy evidence. Overall, the results highlight that \ourmodel is most beneficial when personalization requires resolving implicit intent over long, heterogeneous histories, and it degrades more gracefully as the memory bank becomes noisier.

\section{Additional Experiments}

\subsection{Comparison with Post-Training Baselines}
\begin{table*}[ht]
\centering
\small
\setlength{\tabcolsep}{3pt}
\begin{tabular}{p{2.2cm}|M{1.4cm}M{1.4cm}M{1.4cm}M{1.4cm}M{1.4cm}M{1.4cm}M{1.4cm}|M{1.4cm}}
\toprule
\textbf{Method}  &
\textbf{Recall facts} &
\textbf{Suggest ideas} &
\textbf{Latest prefs} &
\textbf{Prefs evolve} &
\textbf{Update reasons} &
\textbf{Aligned recs} &
\textbf{New Scenarios} &
\textbf{Overall} \\
\midrule
Frozen & 28.93 & 11.39 & - & 44.07 & 61.25 & 35.56 & 15.22 & 34.36 \\
SFT    & 42.15 & \textbf{15.19} & - & 55.93 & \textbf{81.25} & 48.89 & 28.26 & 46.83 \\
PPO    & 51.24 & \textbf{15.19} & - & 60.17 & 75.00 & 44.44 & 36.96 & 49.49 \\
GRPO   & 52.07 & 16.46 & - & 63.56 & 71.25 & 46.67 & 36.96 & 50.31 \\
\midrule
\ourmodel & \textbf{59.50} & 8.86 & - & \textbf{68.64} & \textbf{81.25} & \textbf{62.22} & \textbf{56.52} & \textbf{57.06} \\
\bottomrule
\end{tabular}
\caption{Comparison with different post-training methods on PersonaMem 32K memory corpus data.}
\label{tab:posttrain_comparison_32k}
\end{table*}

Table~\ref{tab:posttrain_comparison_32k} compares \ourmodel with standard post-training baselines trained on the same 300 PersonaMem training data, where the baselines do not perform memory evolution and instead optimize question answering directly. Moving from Frozen to SFT and then to PPO/GRPO yields steady overall improvements (34.36$\rightarrow$46.83$\rightarrow$49.49/50.31), indicating that post-training helps, but the gains are uneven across personalization skills. In contrast, \ourmodel achieves the best overall score (57.06) and leads on most categories, including Recall facts (59.50), Prefs evolve (68.64), Aligned recs (62.22), and New Scenarios (56.52). This pattern is consistent with our design: by explicitly internalizing memory extraction, update, and forgetting into a dedicated memory-evolution mechanism, \ourmodel improves long-horizon preference tracking and generalization beyond what QA-only post-training captures.

\subsection{Evaluation of Preference Retention}

\begin{figure*}[t]
\centering
\includegraphics[width=0.89\linewidth]{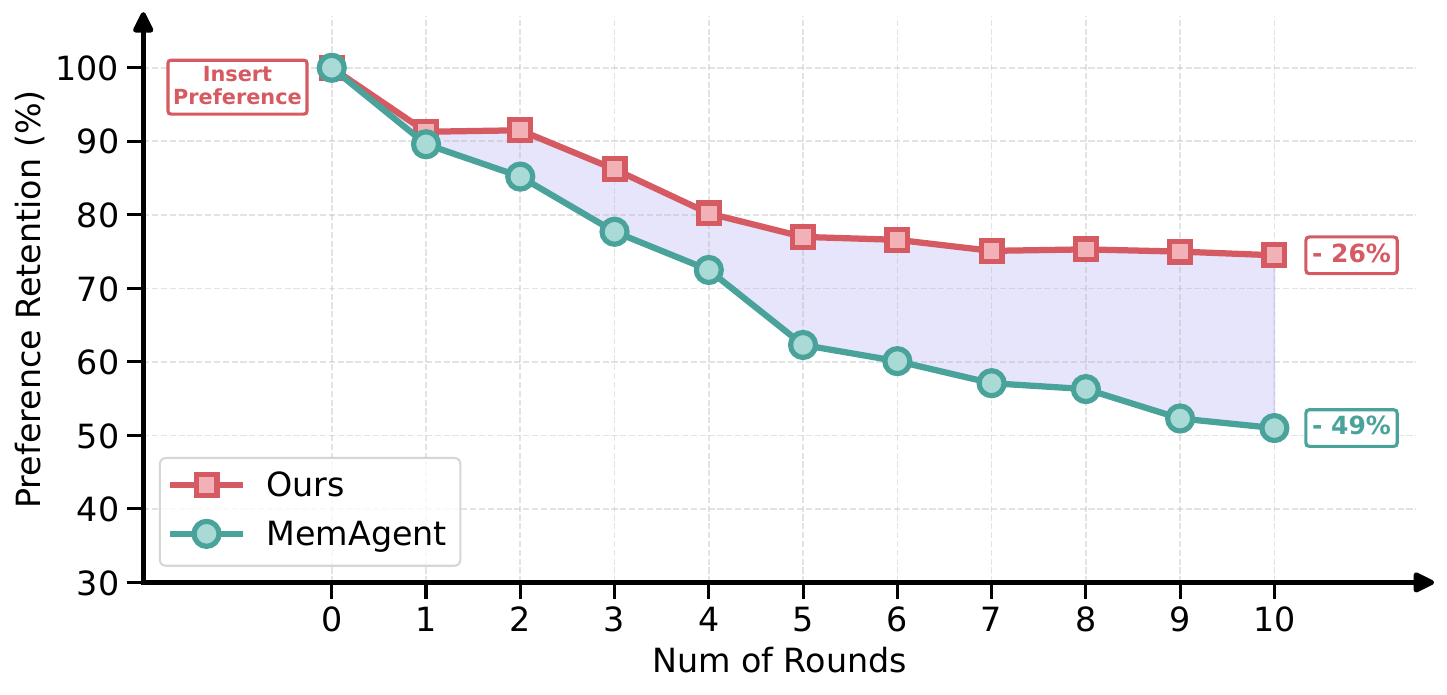}
\caption{Preference retention during multi-round memory evolution on PrefEval (Explicit). We insert a user preference at round 0 and then run memory evolution for subsequent rounds, where each round uses a 4K-token dialogue context. A strong judge model (\texttt{Gemini-2.5-Pro}) verifies whether the preference remains in the memory bank after each round.}
\label{fig:preference_retention_gap}
\end{figure*}
\begin{figure}[!htbp]
\centering
\includegraphics[width=0.89\linewidth]{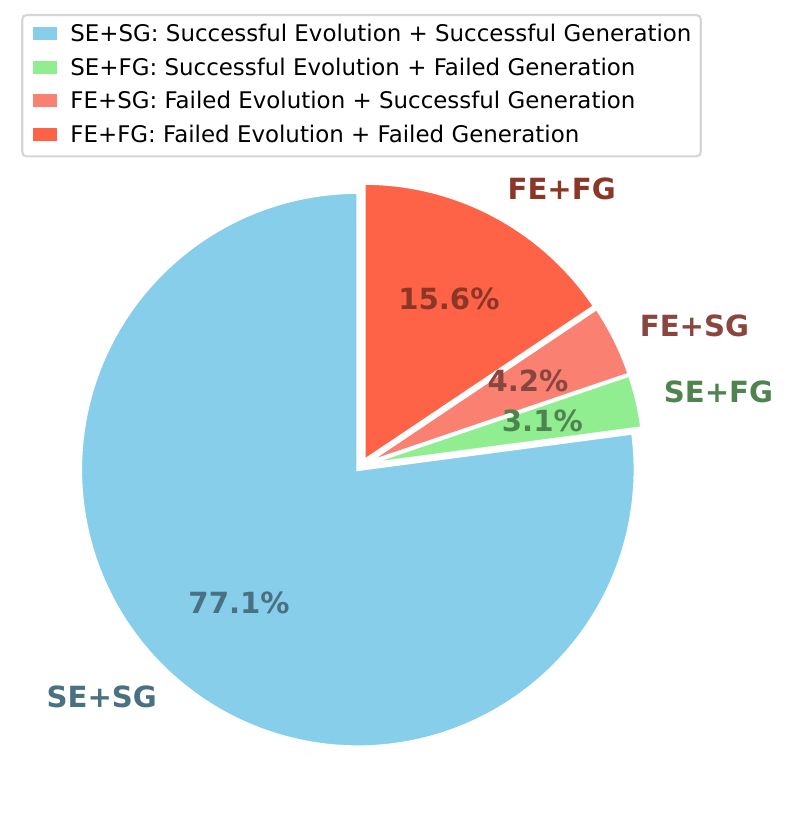}
\caption{Error analysis decomposing failures into the memory evolution stage and the response generation stage. We use the same setup as Figure~\ref{fig:preference_retention_gap}, where successful evolution means the memory bank captures the user preference, and successful generation means the final answer is correct given the evolved memory.}
\label{fig:error-analysis}
\end{figure}

Figure~\ref{fig:preference_retention_gap} directly tests whether our method can preserve useful user preference information inside the memory bank throughout multi-round memory evolution. We use PrefEval (Explicit) because the preference is inserted at the beginning (round 0), which makes retention measurable and avoids ambiguity about when the preference should appear. After inserting the preference, we run multi-round evolution with a fixed 4K-token context per round, and ask \texttt{Gemini-2.5-Pro} to judge whether the memory bank still contains the inserted preference. At round 0, both methods have 100\% retention by construction, after which their retention curves diverge rapidly as rounds increase: MemAgent exhibits a steep and nearly monotonic decay, dropping to roughly 51\% retention by round 10, whereas our method degrades much more slowly and remains around 74\% at round 10. This yields a substantially smaller absolute decrease in retention for our method (about 26\%) compared to MemAgent (about 49\%), and the growing shaded gap indicates that the advantage accumulates over time rather than being a one-off effect. 
Qualitatively, this behavior aligns with our design goal. Specifically, by introducing an induced memory-update guideline to regulate what to keep, refine, or delete, our evolution process is less prone to overwriting or dilution of the initially injected preference under long interaction histories.
In contrast, MemAgent appears more vulnerable to preference drift and forgetting as the number of rounds increases, since later interactions can introduce competing signals and noisy content that interfere with the original preference.

\begin{figure*}[!htbp]
\centering
\includegraphics[width=0.48\linewidth]{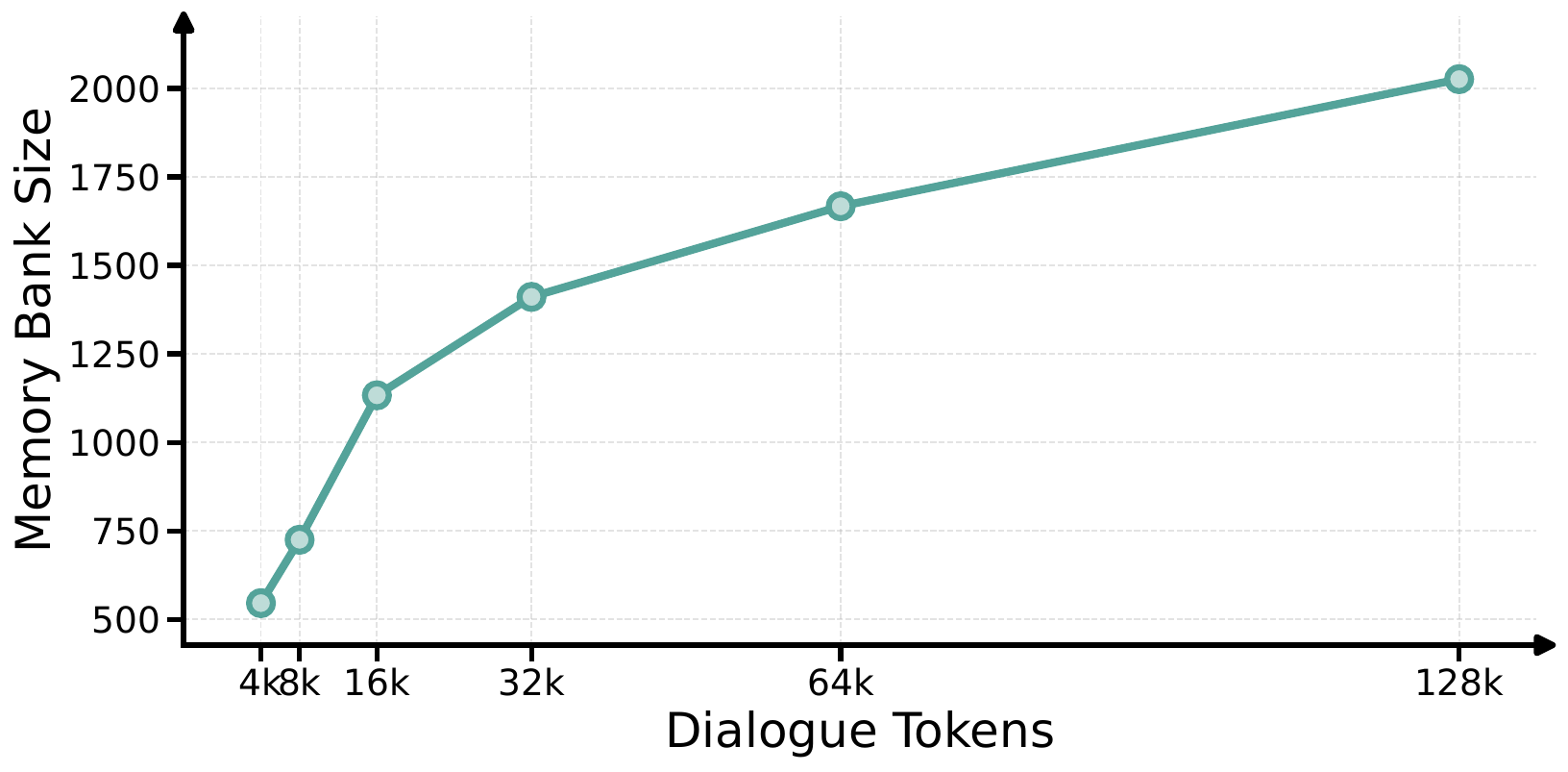}
\includegraphics[width=0.48\linewidth]{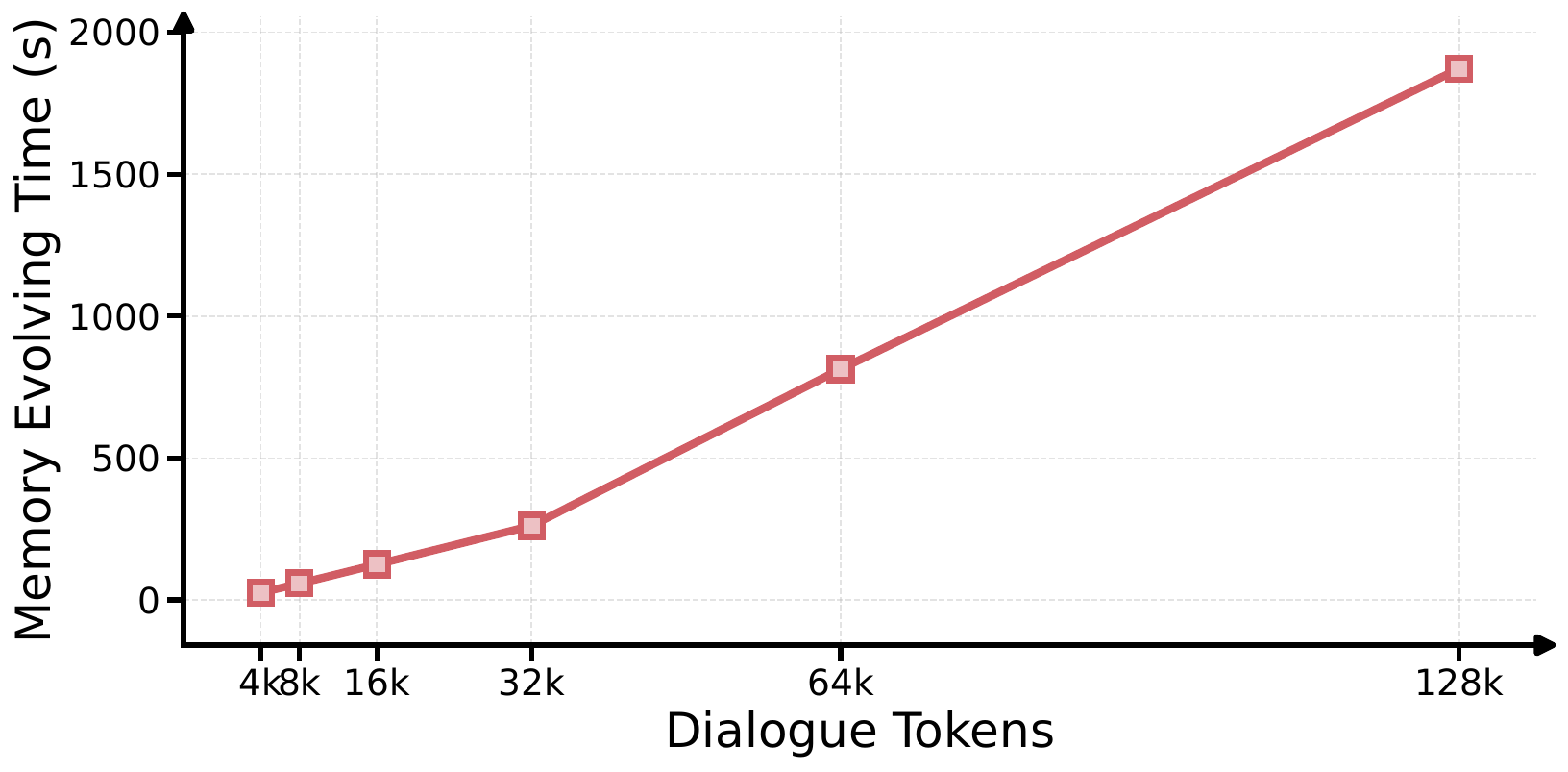}
\caption{Scaling analysis on PersonaMem (128K). We increase the total dialogue tokens (each evolution round processes 4K tokens) and report the resulting memory bank size (left) and memory evolving time (right).}
\label{fig:scaling}
\end{figure*}

\begin{figure*}[t]
    \centering

    \begin{subfigure}[t]{0.42\linewidth}
        \centering
        \includegraphics[width=\linewidth]{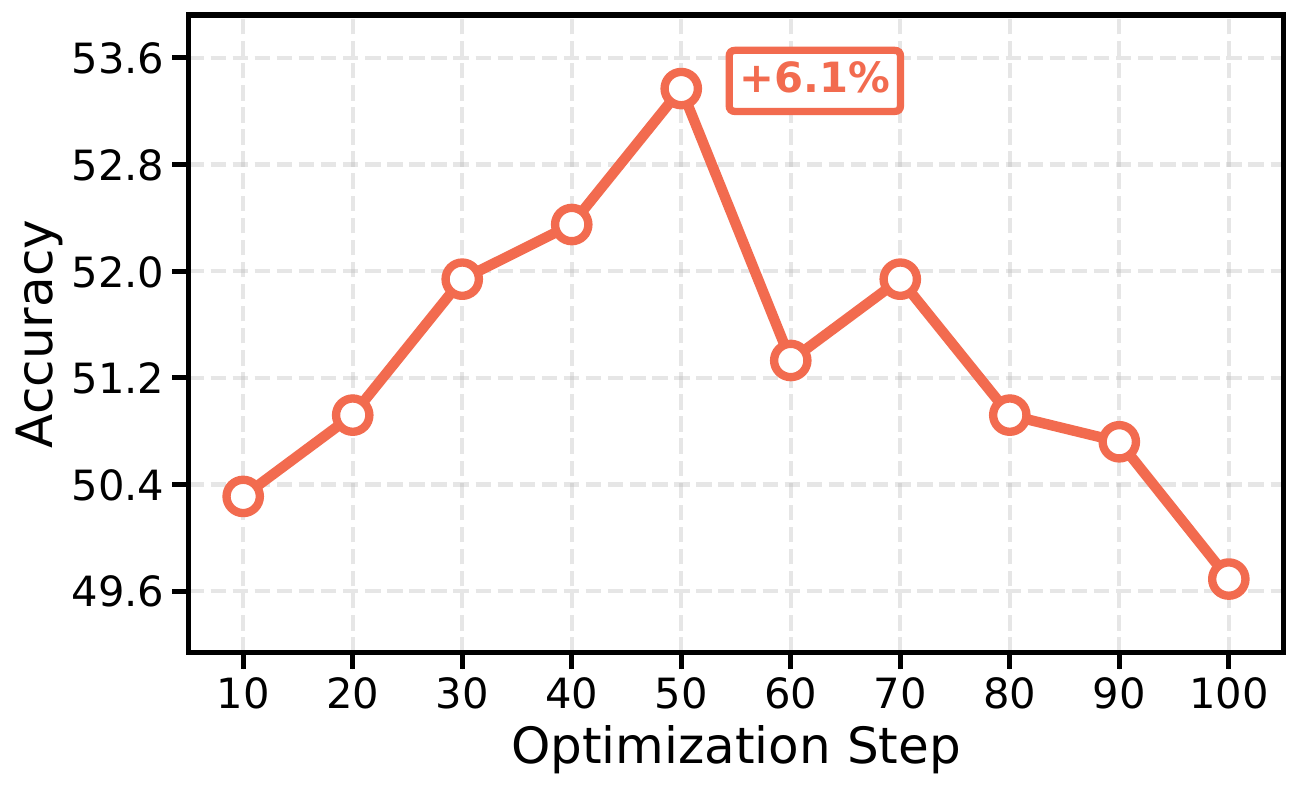}
        \caption{PersonaMem (32K)}
        \label{fig:optim-step-personamem-32k}
    \end{subfigure}
    \begin{subfigure}[t]{0.42\linewidth}
        \centering
        \includegraphics[width=\linewidth]{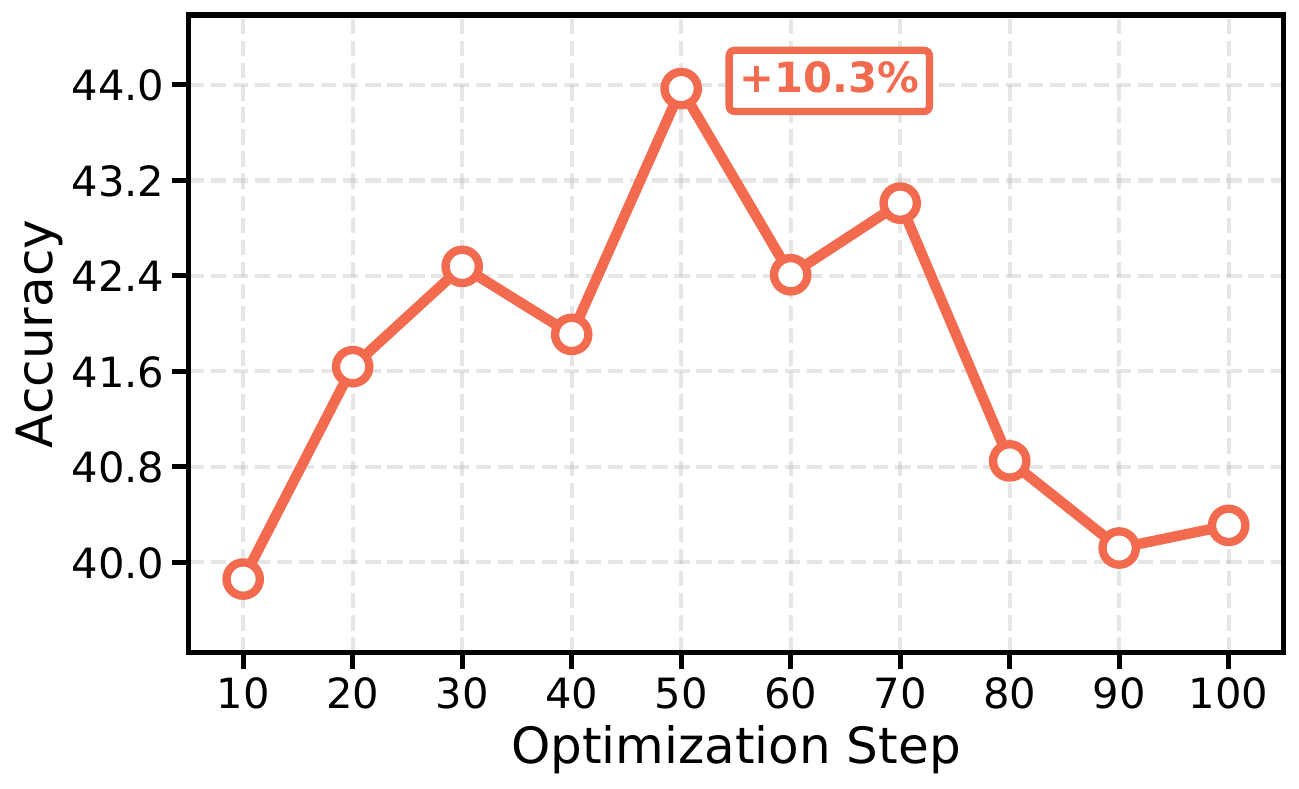}
        \caption{PersonaMem (128K)}
        \label{fig:optim-step-personamem-128k}
    \end{subfigure}

    \vspace{0.6em}

    \begin{subfigure}[t]{0.42\linewidth}
        \centering
        \includegraphics[width=\linewidth]{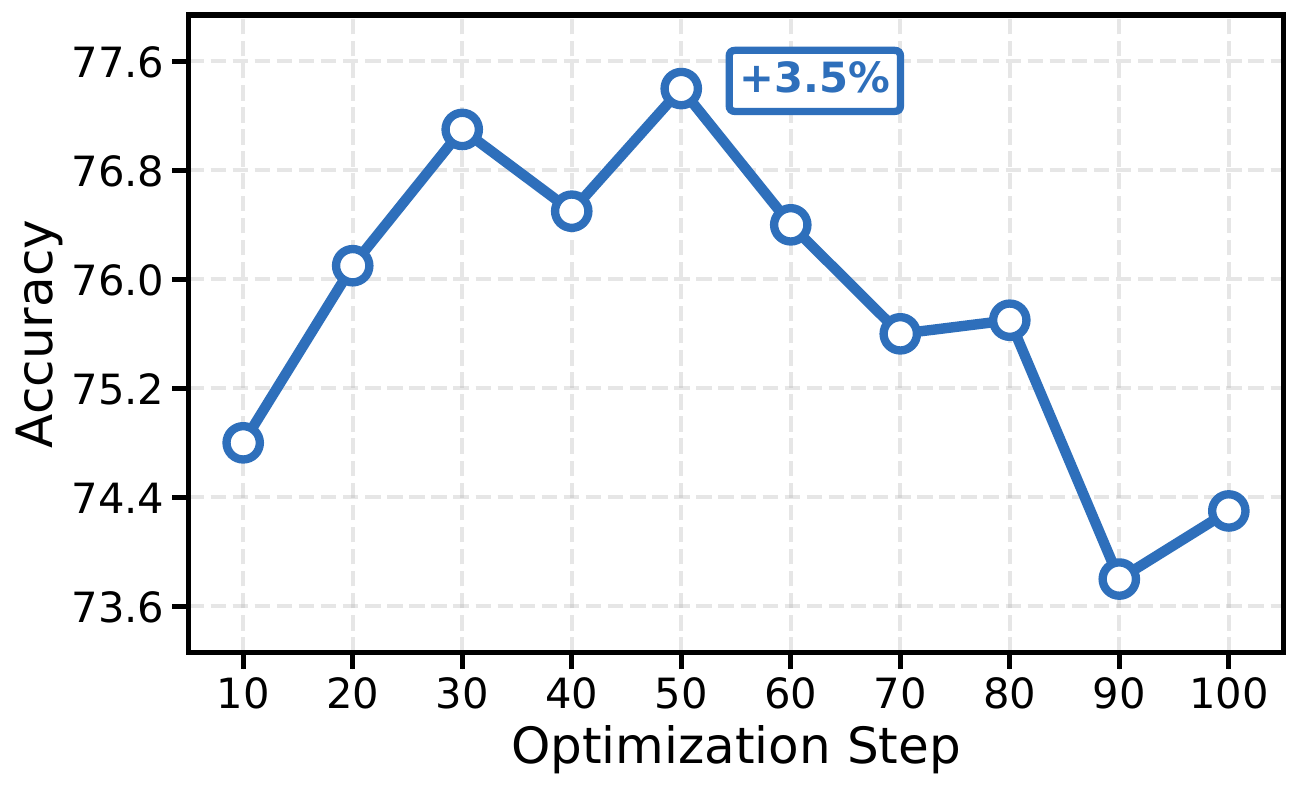}
        \caption{PrefEval (Explicit)}
        \label{fig:optim-step-prefeval-explicit}
    \end{subfigure}
    \begin{subfigure}[t]{0.42\linewidth}
        \centering
        \includegraphics[width=\linewidth]{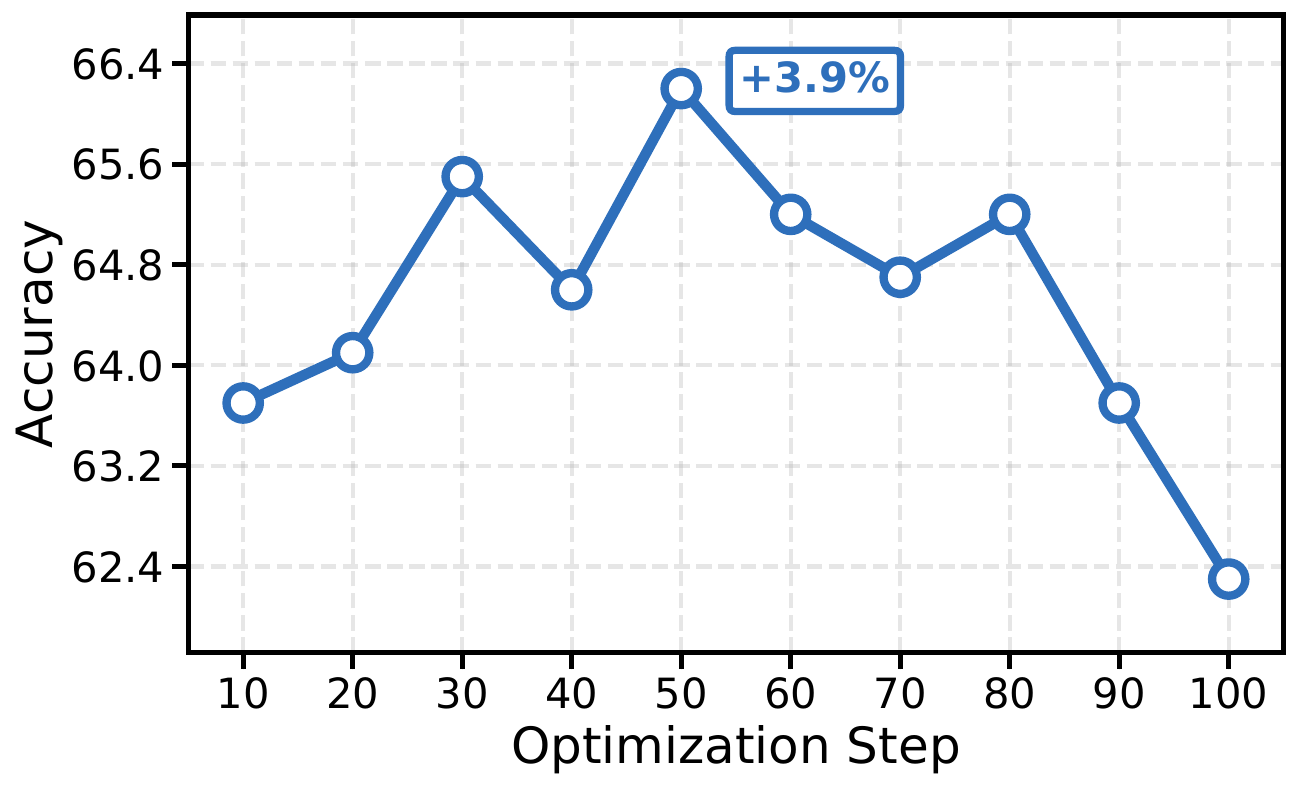}
        \caption{PrefEval (Implicit)}
        \label{fig:optim-step-prefeval-implicit}
    \end{subfigure}
    \caption{Hyperparameter analysis of optimization steps. The relative gains ($+\Delta\%$) are computed w.r.t.\ the performance at step 10.}
    \label{fig:optim-step-hparam}
\end{figure*}

\subsection{Error Analysis}
Figure~\ref{fig:error-analysis} breaks down errors according to whether they originate from the memory evolution stage or the response generation stage under the same preference-retention setup as Figure~\ref{fig:preference_retention_gap}. The dominant category is SE+SG (Successful Evolution + Successful Generation) at 77.1\%, indicating that in most cases the system both captures the user preference in the memory bank and leverages it to produce a correct response. The remaining 22.9\% errors are split across three failure modes: SE+FG accounts for 3.1\%, showing that even when the preference is correctly stored, generation can still fail to use it; FE+SG accounts for 4.2\%, suggesting that correct answers can occasionally be produced despite imperfect preference capture (e.g., the model may rely on residual context cues rather than the memory bank); and FE+FG accounts for 15.6\%, which is the largest failure category and highlights that missed or incorrect preference capture during memory evolution often cascades into downstream response failures. Overall, this decomposition suggests that improving the reliability of the evolve stage, i.e., making it more consistent in capturing and preserving preferences, should yield the largest payoff. The comparatively small SE+FG slice indicates that, although present, generation errors are not the primary bottleneck in this setting.

\subsection{Effect of Training Steps on RL-Based Baselines}

\begin{figure}[t]
    \centering
    \includegraphics[width=1\linewidth]{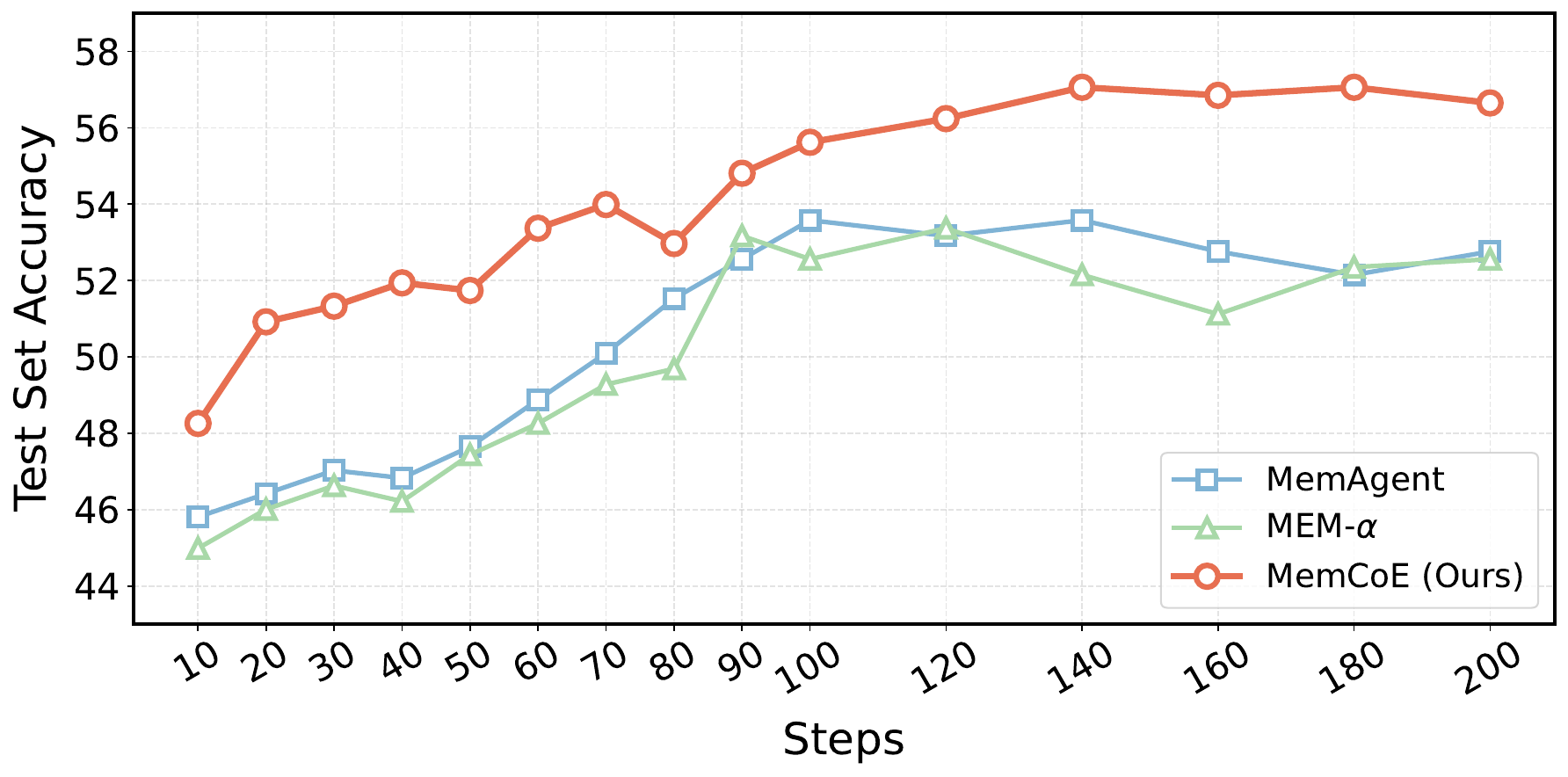}
    \caption{Test set Accuracy of RL-based baselines across RL training steps.}
    \label{fig:training_steps}
\end{figure}

To verify that the performance gap between MemCoE and RL-based baselines is not attributable to insufficient baseline training, we conduct a training-step study under an identical data budget and experimental setting: 300 sampled PersonaMem training examples, batch size 4, and evaluation on PersonaMem-32K, varying only the number of RL update steps up to 200. As shown in Figure~\ref{fig:training_steps}, both MemAgent and MEM-$\alpha$ reach a clear performance plateau around steps 100--140, with MemAgent peaking at 53.58 (steps 100/140) and MEM-$\alpha$ peaking at 53.37 (step 120), after which neither baseline exhibits consistent improvement—confirming that both models have converged within the given budget. Despite this, MemCoE continues to outperform the best checkpoints of both baselines by a substantial margin throughout training, demonstrating that the observed accuracy gap stems primarily from MemCoE's architectural design rather than any insufficiency in baseline training.

\subsection{Scaling Analysis}
Figure~\ref{fig:scaling} analyzes how our memory system scales with longer dialogue context on PersonaMem (128K) using 4K tokens per evolution round. As the dialogue tokens grow from 4K to 128K, the memory bank size increases from roughly 500 to around 2,000, and the curve is sublinear: it grows faster in the short-context regime and then gradually flattens as the history lengthens, which is consistent with consolidating stable information while removing redundant or outdated content to \textbf{keep memory overhead under control}. Meanwhile, the memory evolving time increases smoothly with dialogue length and follows an \textbf{approximately linear trend}, remaining well-behaved across the entire range; this indicates that the computational cost scales predictably with the amount of dialogue processed per round, with the slowly expanding memory bank introducing only a mild additional overhead in longer contexts.

\subsection{Optimization Step Analysis}
Figure~\ref{fig:optim-step-hparam} studies the number of optimization steps used in \moduleone for Memory Guideline Induction. Across all four settings, performance improves from the initial configuration and reaches a clear peak at an intermediate step budget, with the best results achieving relative gains of +6.1\% (PersonaMem 32K), +10.3\% (PersonaMem 128K), +3.5\% (PrefEval Explicit), and +3.9\% (PrefEval Implicit) compared to the 10-step setting. When the step count is too small, the guideline updates are likely under-developed: the aggregated textual gradients have limited opportunity to accumulate recurring error patterns across batches, so the induced guideline remains close to the initial prompt and cannot consistently regulate memory operations. Conversely, when the step count becomes large, the curves show a downward trend after the peak, suggesting diminishing returns and instability: repeated natural-language edits can over-specialize the guideline to feedback from later batches, or amplify small contradictions across textual gradients, which in turn weakens its ability to \textbf{generalize across histories and query types}. Overall, the results indicate that \moduleone benefits from enough iterations to consolidate batch-level critiques into a robust global policy, but requires a moderate step budget to avoid drifting away from broadly useful memory-update principles.

\subsection{Comparison with TextGrad}
\label{subsec:prompt_opt}

To further validate MGI, we compare against TextGrad~\cite{yuksekgonul2025optimizing}, a strong general-purpose prompt optimizer, under the PersonaMem (32K) setting. Results are shown in Table~\ref{tab:prompt_opt}: TextGrad improves over the manual prompt but still lags behind MGI by a substantial margin. This gap highlights that general prompt optimization is insufficient for our memory-evolution setting, where updates must be grounded in long-horizon trajectories. The results support that MGI's trajectory-grounded contrastive signal and batch aggregation are critical for inducing a high-quality memory guideline.

\begin{table}[h]
\centering
\caption{Comparison with prompt optimization methods on PersonaMem (32K). $^*$ indicates statistically significant improvement over the second-best baseline (two-sided $t$-test, $p < 0.05$).}
\label{tab:prompt_opt}
\begin{tabular}{lc}
\toprule
\textbf{Method} & \textbf{PersonaMem (32K)} \\
\midrule
Manual Prompt         & 48.25 {\scriptsize $\pm$ 0.68} \\
TextGrad & 49.83 {\scriptsize $\pm$ 0.83} \\
MemCoE (w/ only MGI)  & \textbf{53.28} {\scriptsize $\pm$ 0.76}$^*$ \\
\bottomrule
\end{tabular}
\end{table}

\lstdefinestyle{prompt}{
    basicstyle=\ttfamily\fontsize{7pt}{8pt}\selectfont,
    frame=none,
    breaklines=true,
    backgroundcolor=\color{white},
    breakatwhitespace=true,
    breakindent=0pt,
    escapeinside={(*@}{@*)},
    numbers=none,
    numbersep=5pt,
    xleftmargin=5pt,
    aboveskip=2pt,
    belowskip=2pt,
}

\tcbset{
  aibox/.style={
    top=10pt,
    colback=white,
    enhanced,
    center,
  }
}
\newtcolorbox{AIbox}[2][]{aibox, title=#2,#1}

\colorlet{PBRedFrame}{red!60}
\colorlet{PBPurpleFrame}{purple!60}
\colorlet{PBOrangeFrame}{orange!60}
\colorlet{PBBlueFrame}{blue!50}

\tcbset{
  aiboxFrameOnly/.style={
    aibox,
    colback=white,      
    colframe=black,     
    boxrule=0.8pt,
    arc=2pt,
    left=6pt,right=6pt,top=6pt,bottom=6pt,
  },
}

\newtcolorbox{AIboxRedFrame}[2][]{aiboxFrameOnly, colframe=PBRedFrame, title=#2,#1}
\newtcolorbox{AIboxPurpleFrame}[2][]{aiboxFrameOnly, colframe=PBPurpleFrame, title=#2,#1}
\newtcolorbox{AIboxOrangeFrame}[2][]{aiboxFrameOnly, colframe=PBOrangeFrame, title=#2,#1}
\newtcolorbox{AIboxBlueFrame}[2][]{aiboxFrameOnly, colframe=PBBlueFrame, title=#2,#1}





\section{Case Study}
Figures~\ref{fig:case_pm1_weekend}--\ref{fig:case_pb2_work} show four representative cases from PersonaMem and PersonaBench. In the two PersonaMem multiple-choice examples (Figures~\ref{fig:case_pm1_weekend} and \ref{fig:case_pm2_fulfill}), \ourmodel selects the ground-truth option (\textbf{(d)} and \textbf{(a)}), whereas all baselines choose different options, indicating that they fail to preserve or exploit the preference-relevant evidence needed for preference-aligned recommendations. In the two PersonaBench factual QA examples (Figures~\ref{fig:case_pb1_age} and \ref{fig:case_pb2_work}), \ourmodel correctly outputs the user’s age (\textbf{39}) and work location (\textbf{University}), while baselines frequently return \emph{Unknown/Not specified} or claim missing information, consistent with information being unavailable at answer time due to memory evolution and/or retrieval failures.

\begin{figure*}[!t]
\begin{AIboxRedFrame}{CASE STUDY (PersonaMem-1): Preference-Aligned Recommendation (MCQ)}
\vspace{1mm}
{\footnotesize
\setlength{\parindent}{0pt}
\setlength{\parskip}{2pt}

\textbf{User question:} I'm planning a weekend getaway and want to try something creatively fulfilling. What would you suggest? \\

\textbf{Question type:} \textit{provide\_preference\_aligned\_recommendations} \\

\textbf{Topic:} \textit{musicRecommendation} \\

\textbf{Options:} \\
\textbf{(a)} How about spending the weekend learning traditional island cooking techniques or delving into the art of Polynesian tattoo design? Imagine mastering the skills of preparing intricate island feasts or crafting culturally rich tattoos on a canvas. This experience would not only honor your heritage but also keep those traditions alive through your newfound expertise. It’s a fantastic way to celebrate your roots and create something truly meaningful! \\
\textbf{(b)} Why not explore the vibrant world of painting by setting up an easel in an art studio or your backyard? Picture yourself dipping brushes into vivid colors, creating masterpieces on canvas inspired by your surroundings. This artistic journey allows you to express emotions through visual art and discover new techniques that can be both relaxing and rewarding, offering a delightful escape from the ordinary. \\
\textbf{(c)} Consider indulging in the art of storytelling by crafting a compelling narrative or writing poetry in a cozy nook. Envision weaving intricate tales or rhythmic verses that captivate the mind and soul, drawing inspiration from your own experiences or the world around you. This literary venture not only sharpens your writing skills but also provides a channel for introspection and endless creativity. \\
\textbf{(d)} How about diving into a soundscape adventure by capturing the symphony of nature in an enchanting forest or by a tranquil lake? Imagine blending the serene rustling of leaves, the melodic rush of water streams, and the rhythmic patter of rain to craft a unique auditory tapestry. This will not only ignite your creativity but also allow you to relive those serene landscapes through your sound engineering skills. It’s a perfect way to reconnect with nature's music and create something truly spectacular! \\

\textbf{Correct answer:} \textbf{(d)} \\
\textbf{Predictions:} \textbf{Ours: (d)}; MemAgent: (a); Mem-$\alpha$: (a); A-Mem: (a); LightMem: (a); Mem0: (c).
}
\end{AIboxRedFrame}
\vspace{-1em}
\caption{PersonaMem case study (MCQ). \textbf{Our method} matches the ground-truth option \textbf{(d)}, while all baselines select different options.}
\label{fig:case_pm1_weekend}
\end{figure*}

\begin{figure*}[!t]
\begin{AIboxRedFrame}{CASE STUDY (PersonaMem-2): Suggesting New Ideas for Music Expression (MCQ)}
\vspace{1mm}
{\footnotesize
\setlength{\parindent}{0pt}
\setlength{\parskip}{2pt}

\textbf{User question:} How can I find a more fulfilling way to express my love for music? \\

\textbf{Question type:} \textit{suggest\_new\_ideas} \\

\textbf{Topic:} \textit{musicRecommendation} \\

\textbf{Options:} \\
\textbf{(a)} You might consider exploring different avenues like writing about your musical journey or experimenting with performing live in settings that inspire you. Also, giving yourself the freedom to simply enjoy music without external pressures could rekindle your passion. \\
\textbf{(b)} Exploring sound engineering might offer a fulfilling way to express your love for music. Like one user who was inspired by a chance meeting with an audio engineer at a festival, you could dive into the world of creating digital music remixes. By blending different influences and styles, you may find inspiration in exploring the nuances of sound capturing and creating unique auditory landscapes that surpass traditional music boundaries. \\
\textbf{(c)} Collaborating with others who share your musical interests can also be a rewarding path. A user found fulfillment through working with musicians from diverse backgrounds, mixing traditional and electronic elements to expand their creative horizons. Such group efforts can enhance not only your musical explorations but also build lasting relationships, as you contribute unique perspectives to a collective musical vision. \\
\textbf{(d)} Consider getting involved in music criticism by writing album reviews. As highlighted by a diligent user after attending a workshop, reviews can be instrumental in shaping listener perspectives and enhancing understanding. By delving deeper into the contexts and intentions behind albums, you can enrich your musical experience and articulate your insights, potentially helping others to connect more deeply with the music. \\

\textbf{Correct answer:} \textbf{(a)} \\
\textbf{Predictions:} \textbf{Ours: (a)}; MemAgent: (c); Mem-$\alpha$: (c); A-Mem: (b); LightMem: (d); Mem0: (c).
}
\end{AIboxRedFrame}
\vspace{-1em}
\caption{PersonaMem case study (MCQ). \textbf{Our method} selects the correct option \textbf{(a)}, whereas each baseline chooses a different option.}
\label{fig:case_pm2_fulfill}
\end{figure*}

\begin{figure*}[!t]
\begin{AIboxOrangeFrame}{CASE STUDY (PersonaBench-1): Basic User Fact (Open-form QA)}
\vspace{1mm}
{\footnotesize
\setlength{\parindent}{0pt}
\setlength{\parskip}{2pt}

\textbf{User question:} At what age am I right now? \\
\textbf{Question type:} \textit{Basic information} \\

\textbf{Correct answer:} \textbf{39} \\
\textbf{Predictions:} \\
\textbf{Ours: 39}; \\
MemAgent: Not specified; \\
Mem-$\alpha$: Not specified; \\
A-Mem: Unknown; \\
LightMem: Information not provided; \\
Mem0: Not specified.
}
\end{AIboxOrangeFrame}
\vspace{-1em}
\caption{PersonaBench case study (open-form QA). \textbf{Our method} outputs the correct age (\textbf{39}); baselines answer with missing/unknown information.}
\label{fig:case_pb1_age}
\end{figure*}

\begin{figure*}[!t]
\begin{AIboxOrangeFrame}{CASE STUDY (PersonaBench-2): Work Location (Open-form QA)}
\vspace{1mm}
{\footnotesize
\setlength{\parindent}{0pt}
\setlength{\parskip}{2pt}

\textbf{User question:} What is the address of my work location? \\
\textbf{Question type:} \textit{Basic information} \\

\textbf{Correct answer:} \textbf{University} \\
\textbf{Predictions:} \\
\textbf{Ours: University}; \\
MemAgent: Unknown;\\
Mem-$\alpha$: Not specified; \\
A-Mem: Not specified; \\
LightMem: The provided information does not include the address of your work location.; \\
Mem0: There is no relevant information provided.
}
\end{AIboxOrangeFrame}
\vspace{-1em}
\caption{PersonaBench case study (open-form QA). \textbf{Our method} recovers the correct work location (\textbf{University}), while baselines report missing or unknown information.}
\label{fig:case_pb2_work}
\end{figure*}

\section{Prompts}

\subsection{Meta Prompt for Guideline Optimization}
We implement a three-stage meta-prompt pipeline to optimize the guideline prompt used for memory evolution. As shown in Figure~\ref{prompt:template_loss}, \texttt{TEMPLATE\_LOSS} performs a contrastive diagnosis by comparing a \texttt{correct\_sample} and a \texttt{wrong\_sample}, identifying why the correct update succeeds, why the incorrect update fails, and what systematic issues exist in \texttt{template\_evolve}. Based on multiple such analyses, Figure~\ref{prompt:template_aggr} uses \texttt{TEMPLATE\_AGGR} to aggregate diverse feedback into a single coherent summary, resolving inconsistencies and retaining the most consistent actionable points. Figure~\ref{prompt:template_optim} uses \texttt{TEMPLATE\_OPTIM} to revise \texttt{template\_evolve} by following the aggregated feedback while preserving the required placeholders, producing an instruction prompt for guideline optimization.

\begin{figure*}[!ht] 
\begin{AIboxBlueFrame}{TEMPLATE\_LOSS: Contrastive Feedback for Memory Update}
\vspace{1mm}
{\footnotesize
\setlength{\parindent}{0pt}
\setlength{\parskip}{2pt}

\textbf{TEMPLATE\_LOSS} = """ \\
Below are two examples of memory updates: one labeled as \texttt{correct\_case} and the other as \texttt{wrong\_case}. \\
These examples illustrate the process of updating memory blocks based on a user's question using \texttt{template\_evolve} \\
by querying a Large Language Model. \\

Your task is to apply \textbf{contrastive learning principles} to thoroughly compare the correct and incorrect samples. \\
Analyze the reasons why the \textbf{correct sample is successful} and identify the factors contributing to the \textbf{failure of the wrong sample}. \\
Critically evaluate the issues present in \texttt{template\_evolve}, and propose how it can better capture \textbf{user preferences}. \\
Provide \textbf{actionable suggestions and feedback} for optimizing the template. \\

\texttt{<correct\_sample>} \\
\textcolor{blue}{\{correct\_sample\}} \\
\texttt{</correct\_sample>} \\

\texttt{<wrong\_sample>} \\
\textcolor{blue}{\{wrong\_sample\}} \\
\texttt{</wrong\_sample>} \\

\textbf{\# Definitions:} \\
\# - \texttt{user\_question\_or\_message}: Represents the question or message provided by the user. \\
\# - \texttt{correct\_answer}: Denotes the correct answer to the given question. \\
\# - \texttt{model\_response}: Indicates the model's response, derived by combining the \texttt{user\_question\_or\_message} with the \texttt{memory\_bank}. \\
\# - \texttt{memory\_bank}: Refers to a storage area that is updated during the memory refinement process. \\

\textbf{\# The template requiring further refinement and optimization is as follows:} \\

\texttt{<template\_evolve>} \\
\textcolor{blue}{\{template\_evolve\}} \\
\texttt{</template\_evolve>} \\

\textbf{\# Please provide suggestions and feedback on how this template can be improved and optimized.} \\
\# Do not directly update the template content; focus solely on providing recommendations. \\
"""
}
\end{AIboxBlueFrame}
\vspace{-1em}
\caption{Meta prompt for generating contrastive feedback by comparing a correct and an incorrect memory-update case, and for diagnosing weaknesses in \texttt{template\_evolve} (TEMPLATE\_LOSS).}
\label{prompt:template_loss}
\end{figure*}


\begin{figure*}[!ht] 
\begin{AIboxBlueFrame}{TEMPLATE\_AGGR: Feedback Aggregation into a Coherent Summary}
\vspace{1mm}
{\footnotesize
\setlength{\parindent}{0pt}
\setlength{\parskip}{2pt}

\textbf{TEMPLATE\_AGGR} = """ \textbf{You are provided with multiple feedback responses from different analyses.} Your task is to \textbf{synthesize these into a single, coherent feedback summary.} Ensure that the final feedback: \\
1. \textbf{Captures the key insights} from each response. \\
2. \textbf{Resolves any conflicting points} by identifying the most consistent and relevant information. \\
3. Provides a \textbf{clear and concise summary} that reflects the overall consensus of the feedback. \\

\textbf{Feedback Responses:} \\
\textcolor{blue}{\{feedback\_responses\}} \\

\textbf{Final Feedback Summary:} \\
"""
}
\end{AIboxBlueFrame}
\vspace{-1em}
\caption{Meta prompt for synthesizing multiple analysis outputs into a single consolidated feedback summary (TEMPLATE\_AGGR).}
\label{prompt:template_aggr}
\end{figure*}


\begin{figure*}[!ht] 
\begin{AIboxBlueFrame}{TEMPLATE\_OPTIM: Template Refinement Under Placeholder Constraints}
\vspace{1mm}
{\footnotesize
\setlength{\parindent}{0pt}
\setlength{\parskip}{2pt}

\textbf{TEMPLATE\_OPTIM} = """ \\
Please refine and optimize the following instruction template \texttt{template\_evolve} based on the provided feedback. \\

\textbf{\# CRITICAL RULES:} \\
\textbf{\# 1. You MUST preserve the following placeholder tokens in the template:} \\
\# \ \ \ For \texttt{template\_evolve\_new}: \textcolor{blue}{\{memory\}}, \textcolor{blue}{\{chunk\}} \\
\# \ \ \ These placeholders are \textbf{REQUIRED} and must appear exactly as shown (with curly braces) in your output. \\
\# \ \ \ If any of these placeholders are missing, the template will fail to work. \\
\textbf{\# 2. Do NOT add any new placeholders.} Only use the placeholders listed above. \\
\# \ \ \ Output the optimized prompt text directly without introducing any additional placeholder tokens. \\

\textbf{\# Below is the template that needs to be revised and improved:} \\

\texttt{<template\_evolve>} \\
\textcolor{blue}{\{template\_evolve\}} \\
\texttt{</template\_evolve>} \\

\textbf{\# Please comprehensively adhere to the feedback provided below to update the template:} \\

\texttt{<feedback>} \\
\textcolor{blue}{\{feedback\}} \\
\texttt{</feedback>} \\

\textbf{\# Place the optimized version of the template within the following tags:} \\

\texttt{<template\_evolve\_new>} \\
\texttt{</template\_evolve\_new>} \\
"""
}
\end{AIboxBlueFrame}
\vspace{-1em}
\caption{Meta prompt for updating \texttt{template\_evolve} using aggregated feedback while strictly preserving required placeholders and forbidding new ones (TEMPLATE\_OPTIM).}
\label{prompt:template_optim}
\end{figure*}


\subsection{Prompt for Memory Evolution and Final Answer Generation}
To enable long-horizon personalization, we first prompt the model to evolve a structured user memory profile from newly observed dialogue chunks under evidence-bounded extraction and conflict-aware consolidation (Fig.~\ref{prompt:template_evolve_step50}). Notably, this prompt is progressively optimized rather than manually fixed: starting from a generic read \& update instruction, it gradually evolves into a more constrained memory policy that emphasizes recency and usefulness, and finally enforces evidence-bounded updates, explicit conflict resolution, and selective exclusion of privacy-sensitive or unsupported content. This evolution is motivated by two recurring failure modes observed during optimization: unresolved conflicts can lead to inconsistent memory, while over-collection of one-off details can make the profile noisy and less useful for personalization.

For fair comparison, we then adopt shared final-answer prompts across all compared methods. As shown in Fig.~\ref{prompt:final_mcq}, for multiple-choice benchmarks (PersonaMem, and PrefEval), the template instructs the model to select the most appropriate option based on user preferences in memory and to output only the option letter, enforcing a strict and comparable output format. For PersonaBench, which requires open-form answers, we use a separate template (Fig.~\ref{prompt:final_open}) that constrains the output to only the name(s) of the relevant entity/entities and explicitly avoids any additional explanation, thereby standardizing response granularity and ensuring fair evaluation under identical prompting conditions.


\begin{figure*}[!ht]
\begin{AIboxPurpleFrame}{TEMPLATE\_EVOLVE\_STEP50: Evidence-Bound Memory Update from a New Chunk}
\vspace{1mm}
{\footnotesize
\setlength{\parindent}{0pt}
\setlength{\parskip}{2pt}

\textbf{TEMPLATE\_EVOLVE\_STEP50} = """ \\

\textbf{You are given a question with options, some new memory, and previous user memory.} Read the new section and update the user memory by prioritizing \textbf{recent and relevant information} that aligns with the user's \textbf{preferences and experiences}. \\

\textbf{<memory>} \textcolor{blue}{\{memory\}} \textbf{</memory>} \\

\textbf{<section>} \textcolor{blue}{\{chunk\}} \textbf{</section>} \\

\textbf{Update rules:} \\
- \textbf{Evidence-bound:} Extract candidate memory items from chunk. Every stored item must be \textbf{directly supported} by chunk. \\
- \textbf{Relevance \& stability:} Store \textbf{long-term preferences}, \textbf{stable facts}, \textbf{recurring habits}, \textbf{long-term goals}, and \textbf{interaction preferences}. Do \textbf{not} store one-off details (e.g., transient locations, momentary moods) unless explicitly stated as long-term. \\
- \textbf{Conflict handling:} If new info contradicts existing memory, prefer the \textbf{most recent supported} info as active, and mark the older one as \textbf{deprecated} with a short reason. \\
- \textbf{Privacy:} Do \textbf{NOT} store highly sensitive or uniquely identifying data (exact address, account credentials, financial/medical specifics, etc.). \\
- \textbf{No domain bias:} Do not assume any specific hobbies or interests unless stated in chunk. \\
- If chunk contains explicit user corrections/ratings about previous outputs, store them under \textbf{"interaction\_feedback"}. Otherwise, do not create feedback entries. \\
- Merge into a \textbf{structured memory profile}. Keep it concise, non-redundant, and internally consistent. \\

"""
}
\end{AIboxPurpleFrame}
\vspace{-1em}
\caption{Meta prompt for updating the user memory profile from a newly observed dialogue chunk with evidence-bounded extraction and conflict-aware consolidation (TEMPLATE\_EVOLVE\_STEP50).}
\label{prompt:template_evolve_step50}
\end{figure*}

\begin{figure*}[!ht]
\begin{AIboxPurpleFrame}{TEMPLATE\_FINAL (MCQ): PersonaMem  / PrefEval}
\vspace{1mm}
{\footnotesize
\setlength{\parindent}{0pt}
\setlength{\parskip}{2pt}

\textbf{TEMPLATE\_FINAL} = """ \textbf{You are presented with a question and its corresponding options.} Find the most appropriate option to the question based on \textbf{user preference in memory} and give your final answer (a), (b), (c), or (d). Put the answer in \textbf{\textbackslash boxed\{\}}. \\

\textbf{<question>} \textcolor{blue}{\{question\}} \textbf{</question>} \\

\textbf{<options>} \textcolor{blue}{\{options\}} \textbf{</options>} \\

\textbf{<memory>} \textcolor{blue}{\{memory\}} \textbf{</memory>} \\

\textbf{Your answer:} \\
"""
}
\end{AIboxPurpleFrame}
\vspace{-1em}
\caption{Shared prompt for final answer generation on multiple-choice benchmarks (PersonaMem, and PrefEval).}
\label{prompt:final_mcq}
\end{figure*}

\begin{figure*}[!ht]
\begin{AIboxPurpleFrame}{TEMPLATE\_FINAL\_OPEN (Open QA): PersonaBench}
\vspace{1mm}
{\footnotesize
\setlength{\parindent}{0pt}
\setlength{\parskip}{2pt}

\textbf{TEMPLATE\_FINAL\_OPEN} = """ \textbf{You are provided with the following relevant information about a user:} \\

\textbf{<memory>} \textcolor{blue}{\{memory\}} \textbf{</memory>} \\

Answer the question below as \textbf{directly and concisely as possible}, using only the \textbf{name(s) of the relevant entity or entities}. \textbf{Avoid adding any extra words or explanations.} \\

\textbf{<question>} \textcolor{blue}{\{question\}} \textbf{</question>} \\

\textbf{Your answer:} \\
"""
}
\end{AIboxPurpleFrame}
\vspace{-1em}
\caption{Shared prompt for final answer generation on PersonaBench, constraining outputs to only the relevant entity name(s).}
\label{prompt:final_open}
\end{figure*}

\end{document}